\pdfoutput=1
\PassOptionsToPackage{table, xcdraw}{xcolor}
\documentclass[11pt]{article}

\usepackage[final]{acl}

\usepackage{times}
\usepackage{latexsym}
\usepackage{placeins}
\usepackage{pdfpages}
\usepackage{algorithm}
\usepackage{algpseudocode}
\usepackage{amsmath}
\usepackage{booktabs}
\usepackage{multirow}
\usepackage{csquotes}
\usepackage{booktabs}
\usepackage{tcolorbox}
\tcbuselibrary{breakable}
\usepackage{fontawesome5}
\usepackage{CJKutf8}
\usepackage{tabularx}
\usepackage{makecell}
\usepackage{dirtytalk}

\usepackage[T1]{fontenc}

\usepackage[utf8]{inputenc}

\usepackage{microtype}

\usepackage{inconsolata}

\usepackage{graphicx}

\usepackage[final,commandnameprefix=always]{changes}

\title{ViDove: A Translation Agent System with Multimodal Context and Memory-Augmented Reasoning}

\author{
 \textbf{Yichen Lu\textsuperscript{1,2}\thanks{Equal Contribution}\thanks{Project Lead}},
 \textbf{Wei Dai\textsuperscript{1,4}\footnote[1]{}\footnote[2]{}},
 \textbf{Jiaen Liu\textsuperscript{1,8}\footnote[1]{}\footnote[2]{}},
 \textbf{Ching Wing Kwok \textsuperscript{1,3}\footnote[1]{}},
\\
 \textbf{Zongheng Wu\textsuperscript{1,5}\footnote[1]{}},
 \textbf{Xudong Xiao\textsuperscript{1,7}},
 \textbf{Ao Sun\textsuperscript{9}},
 \textbf{Sheng Fu\textsuperscript{1}},
\\
 \textbf{Jianyuan Zhan \textsuperscript{7}},
 \textbf{Yian Wang \textsuperscript{6}},
 \textbf{Takatomo Saito\textsuperscript{8}},
 \textbf{Sicheng Lai \textsuperscript{1}\footnote[2]{}},
\\
 \small{\textsuperscript{1}Pigeon AI,
 \textsuperscript{2}Carnegie Mellon University,
 \textsuperscript{3}Fudan University,}
\\ 
 \small{\textsuperscript{4}University of California San Diego,
 \textsuperscript{5}University of Toronto,
 \textsuperscript{6}University of California Irvine,}
\\
 \small{\textsuperscript{7}University of Illinois Urbana-Champaign,
 \textsuperscript{8}Institute of Science Tokyo,}
\\
 \small{\textsuperscript{9}Hong Kong University of Science and Technology}
\\
}

\begin{document}

\maketitle
\begin{abstract}
LLM-based translation agents have achieved highly human-like translation results and are capable of handling longer and more complex contexts with greater efficiency. However, they are typically limited to text-only inputs. In this paper, we introduce \textbf{ViDove}, a translation agent system designed for multimodal input. Inspired by the workflow of human translators, ViDove leverages visual and contextual background information to enhance the translation process. Additionally, we integrate a multimodal memory system and long-short term memory modules enriched with domain-specific knowledge, enabling the agent to perform more accurately and adaptively in real-world scenarios. As a result, ViDove achieves significantly higher translation quality in both subtitle generation and general translation tasks, with a 28\% improvement in BLEU scores and a 15\% improvement in SubER compared to previous state-of-the-art baselines. Moreover, we introduce \textbf{DoveBench}, a new benchmark for long-form automatic video subtitling and translation, featuring 17 hours of high-quality, human-annotated data. Our code is available \href{https://github.com/pigeonai-org/ViDove}{here.}

\end{abstract}

\section{Introduction}
Recent advances in Large Language Models (LLMs) have demonstrated remarkable capabilities in Machine Translation (MT) tasks ~\cite{robinson-etal-2023-chatgpt,gao2023designtranslationpromptschatgpt, xu2024contrastivepreferenceoptimizationpushing, zhu2024multilingualmachinetranslationlarge}. The integration of autonomous agent frameworks with LLM-based MT has shown promising results in enhancing translation capabilities, pushing modern translation systems closer to human-level professional performance ~\cite{wu-etal-2024-transagents, wang2025deltaonlinedocumentleveltranslation, guo2024agentsimtagentassistedsimultaneousmachine,peter2024multi}. Through the incorporation of long-short memory system and multi-agent strategies, these approaches have achieved significant improvements in both translation quality and efficiency, enabling LLM-based MT to handle document-level translation effectively~\cite{wang2025deltaonlinedocumentleveltranslation}.

\begin{figure*}[ht!]
\centering
\includegraphics[width=0.99\textwidth]{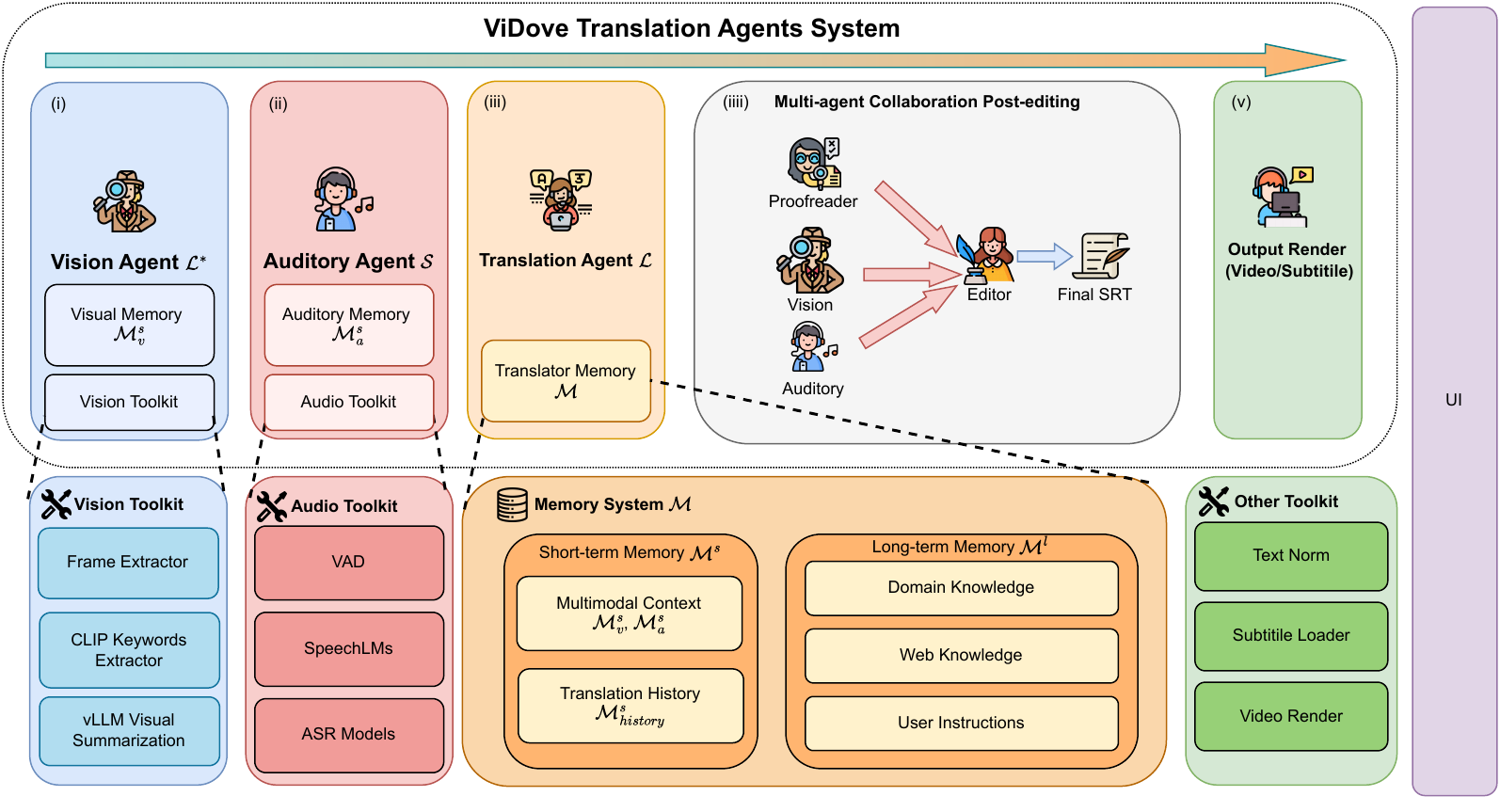} \\
\caption{\textbf{Architecture of the ViDove Translation Agents System.} The system consists of five modules: (i) Vision Agent $\mathcal{L}^*$ and (ii) Auditory Agent $\mathcal{S}$ extract multimodal cues; (iii) Translation Agent $\mathcal{L}$ utilizes memory $\mathcal{M} = \{\mathcal{M}^s, \mathcal{M}^l\}$ for context-aware translation; (iv) a multi-agent post-editing module refines the output via collaboration; (v) Output Render generates final subtitles and video.}
\label{fig:arch}
\end{figure*}

Professional human translators often rely on more than just text to ensure accurate translations. For example, in a cooking video, “fold” could mean combining ingredients or folding a napkin—visual cues like stirring motions and ingredients clarify meaning, while audio tone and emphasis convey intent and emotion~\cite{sulubacak2019multimodalmachinetranslationvisuals, shen2024surveymultimodalmachinetranslation, villacueva2025cammtbenchmarkingculturallyaware}. While some Multimodal Machine Translation(MMT) studies incorporate limited visual or audio inputs, they typically cannot handle document-level translation or take entire video as input\cite{lu2025multimodalmachinetranslationvisual, lv2025topicvdtopicbaseddatasetvideoguided}. However, most existing LLM-based MT systems focus solely on textual input, missing out on these valuable contextual signals.

To close the gap between LLM-based translation and professional human performance, we present \textbf{ViDove}, a multimodal translation agent that integrates visual, audio, and textual inputs. Built on Retrieval-Augmented Generation (RAG)\cite{arslan2024survey, abootorabi2025askmodalitycomprehensivesurvey, zhai2024selfadaptivemultimodalretrievalaugmentedgeneration} and recent Multimodal LLMs(MLLMs) advances\cite{lu2024syneslmunifiedapproachaudiovisual, xu2025qwen25omnitechnicalreport, lu-etal-2024-fastadasp, chu2024qwen2audiotechnicalreport}, ViDove uses a memory system for domain-specific knowledge, multimodal context, and instruction customization~\cite{long2024generativemultimodalknowledgeretrieval, ding2024pdfmvqadatasetmultimodalinformation}. Specialized agents handle these inputs to produce more accurate, human-like translations and subtitles. ViDove achieves state-of-the-art results, with a \textbf{28\%} BLEU and \textbf{15\%} SubER improvement over baselines. We also introduce \textbf{DoveBench}, a video automatic subtitling and translation benchmark with 17 hours of high-quality human-annotated subtitles to support future research.

The key innovations of our work include:

\begin{itemize}
    \item \textbf{Multimodal Multi-Agent Collaboration.} A modular translation agent system that simulates human translator workflows by integrating audio, visual, and textual modalities through specialized agents and their interactions, achieving performance comparable to or better than existing baseline systems.
    \item \textbf{Memory-Augmented Reasoning.} A long-short term memory system for managing multimodal and domain-specific knowledge during translation.
    \item \textbf{DoveBench.} A long-form video automatic subtitling and translation benchmark that reflects real-world subtitling challenges.
\end{itemize}

\section{Related Works}

ViDove is a multimodal translation agent framework that enables cooperative translation through multimodal grounding and memory-guided reasoning. Our work builds on two main areas: multimodal machine translation systems and multi-agent, whose integration remains underexplored.

\noindent\textbf{Machine Translation:} Traditional MT systems~\cite{wu2016googlesneuralmachinetranslation, nllbteam2022languageleftbehindscaling} perform well at the sentence level but struggle with limited contextual cues, particularly in multimodal scenarios. To address this, prior studies~\cite{li-etal-2022-vision, zuo-etal-2023-incorporating, Lin_2020, lan2023exploringbettertextimage} have introduced visual grounding, yet remain constrained to sentence-level tasks with limited context handling. LLM-based MT~\cite{robinson-etal-2023-chatgpt, hendy2023goodgptmodelsmachine, he-etal-2024-exploring, jiao2023chatgptgoodtranslatoryes} achieves strong performance by leveraging longer context, but typically treats LLMs as black-box translators rather than reasoning agents with human-like interpretive capabilities.

\noindent \textbf{RAG for Machine Translation:} Recent studies have increasingly leveraged RAG to enhance MT. Some works applied RAG to improve the quality and cultural grounding of MT by utilizing diverse retrieval pipelines, knowledge graphs, and multi-task fine-tuning setups~\cite{bouthors2024retrievingexamplesmemoryretrieval, conia-etal-2024-towards, wang2024retrievalaugmentedmachinetranslationunstructured, anonymous2024rag}. Concurrently, other works have specifically addressed low-resource languages, showing significant gains by augmenting prompts with retrieved bilingual dictionaries and example sentences~\cite{merx2024lowresourcemachinetranslationretrievalaugmented, chang2025enhancinglowresourceminoritylanguage}. ViDove effectively combines the strengths of these approaches by architecting a multi-agent system where RAG serves as the core information storage and exchange protocol.

\noindent\textbf{Multi-Agent Systems and Translation Agent:}
Recent work on multi-agent LLM systems~\cite{li2023camel, MAS-9423979, MAS-ijcai2024p890, MAS-cheng2024exploringlargelanguagemodel, decision-li2023semanticallyalignedtaskdecomposition, decision-NEURIPS2024_f0ebc318, decision-wang2023avalonsgamethoughtsbattle, decision-xu2024languageagentsreinforcementlearning} shows that dividing complex tasks among specialized agents with distinct roles enhances reasoning and decision-making. These agents collaborate through structured coordination, often leveraging tools~\cite{tool-li2023apibankcomprehensivebenchmarktoolaugmented, tool-ruan2023tptulargelanguagemodelbased, tool-wu2023autogenenablingnextgenllm} and memory systems~\cite{DS-Ding_2023, DS-naturecite-key, DS-singhal2023expertlevelmedicalquestionanswering} to maintain context and task knowledge.

This paradigm has recently been explored in machine translation. For example, TransAgents~\cite{wu-etal-2024-transagents} improves translation quality through multi-agent critique, where agents evaluate and refine each other’s outputs. DelTA~\cite{MAS-mt-wang2025deltaonlinedocumentleveltranslation} ensures document-level consistency using memory modules. Both approaches enhance translation quality in domain-specific or long-form scenarios.

However, these existing approaches remain confined to the textual modality and overlook the potential of MLLMs~\cite{lu2024syneslmunifiedapproachaudiovisual,xu2025qwen25omnitechnicalreport,chu2024qwen2audiotechnicalreport} in enhancing translation quality. To bridge this gap and bring machine translation closer to human-like performance, we propose ViDove, a novel framework that leverages recent advances in both LLM-based agent systems and MLLMs.

\section{ViDove}
In this section, we first introduce the characteristics of long-form video subtitle generation and translation, then describe each agent in Fig. \ref{fig:arch} in detail. 

\subsection{Preliminary} For clarity, we define our primary notation as follows: $\mathcal{V}$ denotes the input video, $\mathcal{L}$ is the translator Agent, $\mathcal{L^*}$ represents the visual agent, and $\mathcal{S}$ signifies the auditory agent. Long-short term memory modules are represented by $\mathcal{M}=\{\mathcal{M}^s,\mathcal{M}^l\}$, capturing both short-term and long-term contexts. The prompt list $P$ contains guiding prompts for analysis tasks. Video chunks are represented by $\mathcal{C}_i$, each consisting of visual ($\mathcal{V}_i$) and audio ($\mathcal{A}_i$) components. Transcripts are denoted by $T_i$, with translated transcripts represented by $T^*_i$. The pipeline framework can be formulated as algorithm \ref{algo}.
\begin{algorithm}[!t]
\caption{ViDove}
\begin{algorithmic}
\Require Input video $\mathcal{V}$, Multi-agent translation agent $\mathcal{L}$ as in Algorithm \ref{algo-mt}, Visual Agent $\mathcal{L^*}$, auditory agent $\mathcal{S}$, Long-short term memory $\mathcal{M}=\{\mathcal{M}^s,\mathcal{M}^l\}$, prompt list $P$
\Ensure Translated video, original and translated transcript tuple $(V^*,T^*,T)$
\State \textbf{Initialize}: $\mathcal{M}^s \leftarrow \emptyset$, $\mathcal{M}^l \leftarrow \text{Knowledge Base}$ \Comment{Initialize memory modules}
\State $\{(\mathcal{V}_i,\mathcal{A}_i)\}_{i=1}^{k}\leftarrow\mathcal{C}(\mathcal{V})$ \Comment{Chunk decomposition}
\For{each chunk $(\mathcal{V}_i,\mathcal{A}_i)\in \mathcal{C}$}
    \State $(\mathcal{M}^s_{v}, cue_v) \leftarrow \mathcal{L^*}(\mathcal{V}_i, \mathcal{M}^s_{v}, \mathcal{M}^l_{domain}, p_{analysis}), \quad p_{analysis}\in P$ \Comment{Update visual cues and short-term memory}
    \State $cue_a \leftarrow \mathcal{S}(\mathcal{A}_i)$
    \State $T_i \leftarrow (cue_a,cue_v)$
    \State $(T^*_i, \mathcal{M}^s_{history}) \leftarrow \mathcal{L}(T_i, \mathcal{M}, p_{translation})$
\EndFor
\State $T^*,T \leftarrow \textbf{Multi-agentPostProcess}(T^*,T, \mathcal{M}, p_{pr})$ \Comment{Final post-processing of translations}
\State \Return $(V^*,T^*,T)$
\end{algorithmic}
\label{algo}
\end{algorithm}

\subsection{Long-form Video Automatic Subtitling}
\label{sec:long-form VAS}
Long-form(>10min) video automatic subtitling requires the system to process a video $\mathcal{V}$, which includes a synchronized audio stream $\mathcal{A}$. Unlike typical sentence-level translation, this task operates at the document level and demands precise alignment of translated subtitles with the corresponding timestamps. Moreover, in contrast to standard document-level MT, the input does not contain any original text. Instead, the system must first perceive the video by “listening” to the audio—and, when necessary, “observing” the visual content—to transcribe the source language before translating it. These multimodal and multi-stage requirements—speech recognition, visual grounding, and context-aware translation—make the task considerably more complex than traditional MT, multimodal MT, or document-level MT (DocMT). Addressing this challenge calls for a holistic, agent-based approach capable of perception, reasoning, and translation.

\subsection{Auditory Agent}

In this section, we describe the workflow of ViDove’s auditory agent, which provides audio-based contextual information and enriched transcriptions to support the Translation Agent.

\noindent\textbf{Step 1: Chunk Splitting and Timestamp Extraction.}
We use Pyannote~\cite{Plaquet23} to segment the input video $\mathcal{V}$ into $k$ chunks, $\mathcal{C}$, based on speaker activity. Each chunk $\mathcal{C}_i$ contains a corresponding set of video frames $\mathcal{V}_i$ and an audio sequence $\mathcal{A}_i$.

\noindent\textbf{Step 2: Auditory Information Extraction.}
ViDove integrates SOTA single-task models to extract key auditory features: speech transcription~\cite{radford2022robustspeechrecognitionlargescale}, background audio event detection~\cite{chen2024eatselfsupervisedpretrainingefficient, chen2022beatsaudiopretrainingacoustic}, and speaker emotion recognition~\cite{ma2023emotion2vecselfsupervisedpretrainingspeech}. In addition to these, we incorporate recent Speech Language Models(SpeechLMs)~\cite{tang2024salmonngenerichearingabilities, chu2024qwen2audiotechnicalreport}, which offer a unified approach for extracting rich audio cues.
For each chunk, the extracted auditory information, denoted as $cue_a$, is stored in the multimodal contextual memory $\mathcal{M}^s_{a} \in \mathcal{M}^s_{multimodal}$ to support downstream translation.

\noindent\textbf{Step 3: Audio Transcription and Timestamp Refinement.}
ViDove’s auditory agent uses either a SpeechLM or an ASR model to transcribe the audio sequence $\mathcal{A}i$, leveraging the multimodal contextual memory $[\mathcal{M}^s_{a}, \mathcal{M}^s_{v}] \in \mathcal{M}^s_{multimodal}$. This fusion enables enhanced transcription through keyword injection and agent-based reasoning.

\subsection{Vision Agent}
ViDove implements a visual agent to gather informative visual cues from video input $\mathcal{V}$. These cues are used to enhance speech recognition~\cite{lu2024syneslmunifiedapproachaudiovisual,wu2024enhancingaudiovisualspeechrecognition} and are passed through a memory system to support more sophisticated sentence comprehension. This visual cue allows the agent to resolve textually ambiguousness and accurately interpret domain-specific terminology during translation. The ViDove system is designed to be model-agnostic and supports multiple vision-language backends, offering deployment flexibility across local and remote environments.

The pipeline is designed to process pre-segmented video chunks $$\mathcal{V}=\{\mathcal{V}_0,\mathcal{V}_1,...,\mathcal{V}_k\},k=|V|$$, from which key frames are extracted to represent salient visual moments. These frames are then analyzed by the selected vision-language model $\mathcal{L}^*$, providing a high-level semantic understanding $cue_v$ of the visual context.
$$cue_v = \mathcal{L^*}(v_i,\mathcal{M}^s_{vision},M^l_{domain},p_{analysis})$$
Where $\mathcal{M}$ is multi-modal memory for ViDove system.

\subsection{Memory system} 
ViDove's memory system, denoted as \(\mathcal{M} = \{\mathcal{M}^s, \mathcal{M}^l\}\), is implemented using LLaMA-Index\cite{Liu_LlamaIndex_2022}. This memory system stores and organizes multimodal information, enabling the system to deliver contextually informed and consistent translations.

\subsubsection{Short-term Memory (\(\mathcal{M}^s\))}
The short-term memory (\(\mathcal{M}^s\)) is tailored to the current video translation task. It holds information specific to each video chunk (\(\mathcal{C}_i\)), which includes visual (\(\mathcal{V}_i\)) and audio (\(\mathcal{A}_i\)) components. Its contents include:

\noindent\textbf{Translation History:} Records of prior translations within the same video, ensuring consistency in terminology and phrasing across transcripts (e.g., from \(T_i\) to \(T^*_i\)).

\noindent\textbf{Visual Cues:} Contextual data extracted from \(\mathcal{V}_i\), such as scene descriptions or objects, that helps to disambiguate textual content.

\noindent\textbf{Audio Cues:} Auditory information extracted from \(\mathcal{A}_i\), denoted as \(cue_a\), including transcriptions and other speaker information, stored in the multimodal contextual memory \(\mathcal{M}^s_{a} \in \mathcal{M}^s_{multimodal}\). 

This component provides immediate context, supporting accurate and coherent translations within a single video.

\subsubsection{Long-term Memory (\(\mathcal{M}^l\))}
The long-term memory (\(\mathcal{M}^l\)) is designed for adaptability across multiple translation tasks. It accumulates knowledge over time, storing:

\noindent\textbf{Domain Knowledge:} This type of knowledge captures specialized community language to ensure accurate video translations for a diverse, multilingual audience.

\noindent\textbf{Web Knowledge:} General information sourced from the web, implemented by Tavily\cite{tavily}, offering broader context.

This component enhances ViDove's flexibility, enabling it to handle diverse domains and improve performance over time.  The memory system acts as a centralized repository, providing essential information that supports the translation process by ensuring consistency and contextual relevance.

\begin{table*}[ht]
\resizebox{\textwidth}{!}{\begin{tabular}{@{}l|cccc|cc@{}}
\toprule
                 & \multicolumn{4}{c|}{DoveBench}                                                             & \multicolumn{2}{c}{BigVideo}                \\ \midrule
                 & BLEU($\uparrow$)                 & BLEURT($\uparrow$)               & SubER($\downarrow$)                & SubSONAR($\uparrow$)             & BLEU($\uparrow$)                 & sCOMET($\uparrow$)               \\ \midrule
\rowcolor[HTML]{EFEFEF} 
Gemini-2.5-Flash &      8.11                &           17.21           &           103.46           &      0.31                 &         26.43             & \multicolumn{1}{c}{0.75}                   \\ \midrule
Qwen-2.5-Omni    &     14.60                 &            13.83          &           108.94           &   0.39                    &            10.67          &\multicolumn{1}{c}{0.58}                     \\ \midrule
VideoCaptioner   & \multicolumn{1}{c}{12.65} & \multicolumn{1}{c}{14.62} & \multicolumn{1}{c}{85.75} & \multicolumn{1}{c|}{\textbf{0.41}} & \multicolumn{1}{c}{\textbf{30.36}} & \multicolumn{1}{c}{\textbf{0.75}} \\
Whisper + DelTA  & \multicolumn{1}{c}{18.26} & \multicolumn{1}{c}{12.30} & \multicolumn{1}{c}{86.83} & \multicolumn{1}{c|}{0.28} & \multicolumn{1}{c}{29.09} & \multicolumn{1}{c}{0.69} \\
ViDove           & \multicolumn{1}{c}{\textbf{23.51}} & \multicolumn{1}{c}{\textbf{19.55}} & \multicolumn{1}{c}{\textbf{73.38}} & \multicolumn{1}{c|}{0.39} & \multicolumn{1}{c}{26.05} & \multicolumn{1}{c}{0.73} \\ \bottomrule

\end{tabular}}
\caption{ViDove compared with different baseline system on DoveBench and BigVideo.}
\label{tab:main}
\end{table*}

\subsection{Multi-agent Translation}
ViDove's translation process relies on a multi-agent system featuring three specialized agents---\textit{Translator}, \textit{Proofreader}, and \textit{Editor}---that work together to produce high-quality subtitle translations. These agents collaborate by accessing the unified memory system \(\mathcal{M} = \{\mathcal{M}^s, \mathcal{M}^l\}\), ensuring consistency and context throughout the workflow.

\subsubsection*{Agents and Their Roles}

\noindent\textbf{Translator Agent (\(\mathcal{L}_t\)):} This agent generates the initial translation (\(T^*_i\)). To achieve this, it interacts with the memory system by retrieving \textit{translation history} and \textit{visual cues} from short-term memory (\(\mathcal{M}^s\)) to maintain intra-video consistency, while simultaneously drawing upon \textit{domain knowledge} from long-term memory (\(\mathcal{M}^l\)) to ensure domain-specific accuracy and stylistic alignment from the outset.

\noindent\textbf{Proofreader Agent (\(\mathcal{L}_{pr}\)):} Proofreader agent focuses on refining the initial translation by correcting grammar, style, and terminology. It interacts with memory by accessing \textit{domain knowledge} from long-term memory (\(\mathcal{M}^l\)) to ensure linguistic precision and adherence to specialized terminology, while referencing the \textit{translation history} in short-term memory (\(\mathcal{M}^s\)) to maintain consistency with prior segments. The proofreader agent generates revision suggestions for the editor agent, which makes the final decision on whether to apply them. A sample interaction log is provided in Appendix~\ref{appendix：proofreader}.

\noindent\textbf{Editor Agent (\(\mathcal{L}_{ed}\)):} The editor agent performs the final quality check and applies necessary modifications to ensure the translation quality with the full modality context. It receives suggestions from the proofreader agent and decides whether to adopt them. It also accepts user instructions, enabling more free-form and practical interactions between the user and agents.  To verify contextual accuracy, the editor agent leverages the memory system by retrieving \textit{visual cues}, \textit{audio cues}, and \textit{translation history} from the short-term memory (\(\mathcal{M}^s\)). It also queries \textit{web knowledge} from the long-term memory (\(\mathcal{M}^l\)) to incorporate broader external context and ensure logical consistency.

\section{DoveBench}
To the best of our knowledge, there is currently no standardized open-sourced benchmark for long-form video automatic subtitling (Sec \ref{sec:long-form VAS}). To address this gap, we introduce \textbf{DoveBench}, an open-source benchmark designed specifically for this task. DoveBench contains approximately 17 hours of video data, each annotated with Chinese (ZH) subtitles translated by professional translators. The average video length is around 20 minutes, reflecting typical durations found in real-world scenarios. Detailed statistics of DoveBench are provided in Appendix~\ref{appendix:dovebench}.

\section{Experiments}

In this section, we first describe the evaluation datasets, baseline systems, metrics, and ViDove’s configuration. We then present and analyze experiment results.

\subsection{Datasets and Metrics}
We evaluate long-form video automatic subtitling on DoveBench and MMT on BigVideo~\cite{kang-etal-2023-bigvideo}. To assess translation quality, we use BLEU~\cite{freitag-etal-2020-bleu}, BLEURT~\cite{sellam2020bleurt}, and sCOMET~\cite{rei-etal-2020-comet}. For subtitle quality—capturing both translation accuracy and timestamp alignment—we adopt SubER~\cite{wilken-etal-2022-suber} and SubSONAR~\cite{gaido-et-al-2024-sbaam}, specifically for evaluating on DoveBench.

\subsection{Baselines and ViDove Configuration}
We compare ViDove against four baseline systems. Gemini-2.5-flash~\cite{gemini2.5flash2025} serves as a proprietary baseline, and Qwen-2.5-Omni~\cite{xu2025qwen25omnitechnicalreport} represents an open-source alternative. Both models are single MLLMs capable of processing video input and generating subtitle (SRT) output through carefully designed prompts. For system-level baselines, we include VideoCaptioner~\cite{videocaptioner2024}, an open-source cascaded pipeline for video subtitling, and DelTA~\cite{wang2025deltaonlinedocumentleveltranslation}, a state-of-the-art text-based translation agent. Since DelTA does not support audio or video input natively, we use \texttt{whisper-large-v3}~\cite{radford2022robustspeechrecognitionlargescale} to first transcribe the audio for the system.
For ViDove, we use Gemini-2.5-flash~\cite{gemini2.5flash2025} as the auditory agent, while all other agents are powered by GPT-4o~\cite{openai2024gpt4ocard}. Note that neither the baseline models nor ViDove undergo any additional fine-tuning during the experiments. Detailed prompts for the baseline models and ViDove are provided in Appendix ~\ref{appendix:gemini},  ~\ref{appendix:qwen} and ~\ref{appendix:vidovePrompt}.

\subsection{Experiment Results}

Table~\ref{tab:main} presents the evaluation results of ViDove and baseline systems on both DoveBench and BigVideo.

On DoveBench, ViDove consistently outperforms all baselines across all metrics, achieving the highest BLEU (23.51), BLEURT (19.55), and the lowest SubER (73.38). Compared to the strongest baseline, Whisper + DelTA, ViDove improves BLEU by 28.8\%, BLEURT by 58.9\%, and reduces SubER by 15.5\%. These results indicate that ViDove not only produces more accurate translations but also aligns subtitles more precisely in time. However, despite ViDove’s leading performance, the absolute scores—especially BLEU and SubER—remain relatively modest. This highlights the intrinsic difficulty of long-form video automatic subtitling. To date, no existing system has achieved fully satisfactory results on this task, underscoring its complexity and open research nature.

In contrast, Gemini-2.5-flash and Qwen-2.5-Omni perform poorly on DoveBench, with high SubER values (103.46 and 108.94, respectively) and low BLEU scores. Although we carefully engineered prompts and applied post-processing to ensure fair evaluation, these single-model MLLMs struggle on long-form inputs. Their limited instruction-following capability, combined with a tendency to hallucinate or ignore constraints as input length increases, leads to misaligned, incomplete, or off-topic subtitles—even when the task is clearly specified.

On BigVideo, which focuses on sentence-level MMT, ViDove remains competitive. While it is not specifically optimized for short-form translation tasks, it achieves BLEU (26.05) and sCOMET (0.73) scores close to the best-performing models, such as Gemini-2.5-flash and VideoCaptioner. This demonstrates ViDove’s generalizability and robustness.

\subsection{Ablation Study}

\begin{CJK}{UTF8}{gbsn}
\begin{table}[t]
\centering
\scriptsize  
\renewcommand{\arraystretch}{1.15}
\begin{tabular}{@{}l@{\hspace{5pt}}c@{\hspace{5pt}}c@{\hspace{5pt}}c@{\hspace{5pt}}c@{}}
\toprule
\textbf{Model} & BLEU~($\uparrow$) & SubER~($\downarrow$) & BLEURT~($\uparrow$) \\
\midrule
ViDove (full) & \textbf{15.84} & \textbf{76.26} & 17.11 \\
w/o domain memory & 14.86 & 77.31 & \textbf{17.84} \\
w/o domain memory \& vision & 14.56 & 77.55 & 17.50 \\
w/o Proofreader&13.56&80.76& 16.93\\
\bottomrule
\end{tabular}
\caption{Ablation study of ViDove under single-column setting.}
\label{tab:vidove_ablation_single}
\end{table}
\end{CJK}

\vspace{-2pt}
To assess the contribution of different components in ViDove, we conduct an ablation study by removing modules. Our ablation study is conducted on a subset of DoveBench's StarCraft 2 Category. 
As shown in Table~\ref{tab:vidove_ablation_single}, removing the domain memory significantly reduces BLEU and SubER scores, though BLEURT slightly improves—likely due to more generic paraphrasing. Excluding the proofreader agent causes the sharpest quality drop, highlighting its role in correction and consistency. While the visual module has limited impact on BLEU or BLEURT, it helps the editor correct entity-level terms (e.g., names and objects), improving factual accuracy and user experience beyond what metrics capture.

\section{Conclusion}
In this work, we introduced \textbf{ViDove}, a multimodal translation agent system for long-form video inputs. Our model outperforms the strongest existing baselines by up to 28.8\% in BLEU and 15.5\% in SubER, demonstrating significant improvements in both translation accuracy and subtitle alignment. We also release a new benchmark for the challenging task of long-form video automatic subtitling. Compared to prior work, ViDove offers a more practical and scalable solution for video automatic subtitling and translation.

\section*{Acknowledgments}
We thank the FGA Subtitle Group, MetricSubs, and Star-Pigeon Group for their support in providing high-quality human-annotated datasets. This work was supported in part by funding from Pigeon AI.

\bibliography{acl_latex}

\begin{thebibliography}{83}
\providecommand{\natexlab}[1]{#1}

\bibitem[{Abootorabi et~al.(2025)Abootorabi, Zobeiri, Dehghani, Mohammadkhani, Mohammadi, Ghahroodi, Baghshah, and Asgari}]{abootorabi2025askmodalitycomprehensivesurvey}
Mohammad~Mahdi Abootorabi, Amirhosein Zobeiri, Mahdi Dehghani, Mohammadali Mohammadkhani, Bardia Mohammadi, Omid Ghahroodi, Mahdieh~Soleymani Baghshah, and Ehsaneddin Asgari. 2025.
\newblock \href {https://arxiv.org/abs/2502.08826} {Ask in any modality: A comprehensive survey on multimodal retrieval-augmented generation}.
\newblock \emph{Preprint}, arXiv:2502.08826.

\bibitem[{Anonymous(2024)}]{anonymous2024rag}
Anonymous. 2024.
\newblock \href {https://openreview.net/forum?id=h4NMSEHcwh} {{RAG} picking helps: Retrieval augmented generation for machine translation}.
\newblock In \emph{Submitted to ACL Rolling Review - August 2024}.
\newblock Under review.

\bibitem[{Arslan et~al.(2024)Arslan, Ghanem, Munawar, and Cruz}]{arslan2024survey}
Muhammad Arslan, Hussam Ghanem, Saba Munawar, and Christophe Cruz. 2024.
\newblock A survey on rag with llms.
\newblock \emph{Procedia Computer Science}, 246:3781--3790.

\bibitem[{Bai et~al.(2025)Bai, Chen, Liu, Wang, Ge, Song, Dang, Wang, Wang, Tang, Zhong, Zhu, Yang, Li, Wan, Wang, Ding, Fu, Xu, Ye, Zhang, Xie, Cheng, Zhang, Yang, Xu, and Lin}]{Qwen2.5-VL}
Shuai Bai, Keqin Chen, Xuejing Liu, Jialin Wang, Wenbin Ge, Sibo Song, Kai Dang, Peng Wang, Shijie Wang, Jun Tang, Humen Zhong, Yuanzhi Zhu, Mingkun Yang, Zhaohai Li, Jianqiang Wan, Pengfei Wang, Wei Ding, Zheren Fu, Yiheng Xu, Jiabo Ye, Xi~Zhang, Tianbao Xie, Zesen Cheng, Hang Zhang, Zhibo Yang, Haiyang Xu, and Junyang Lin. 2025.
\newblock Qwen2.5-vl technical report.
\newblock \emph{arXiv preprint arXiv:2502.13923}.

\bibitem[{Bouthors et~al.(2024)Bouthors, Crego, and Yvon}]{bouthors2024retrievingexamplesmemoryretrieval}
Maxime Bouthors, Josep Crego, and Francois Yvon. 2024.
\newblock \href {https://arxiv.org/abs/2404.02835} {Retrieving examples from memory for retrieval augmented neural machine translation: A systematic comparison}.
\newblock \emph{Preprint}, arXiv:2404.02835.

\bibitem[{Chang et~al.(2025)Chang, Li, Lee, and Lee}]{chang2025enhancinglowresourceminoritylanguage}
Chen-Chi Chang, Chong-Fu Li, Chu-Hsuan Lee, and Hung-Shin Lee. 2025.
\newblock \href {https://arxiv.org/abs/2505.10829} {Enhancing low-resource minority language translation with llms and retrieval-augmented generation for cultural nuances}.
\newblock \emph{Preprint}, arXiv:2505.10829.

\bibitem[{Chen et~al.(2022)Chen, Wu, Wang, Liu, Tompkins, Chen, and Wei}]{chen2022beatsaudiopretrainingacoustic}
Sanyuan Chen, Yu~Wu, Chengyi Wang, Shujie Liu, Daniel Tompkins, Zhuo Chen, and Furu Wei. 2022.
\newblock \href {https://arxiv.org/abs/2212.09058} {Beats: Audio pre-training with acoustic tokenizers}.
\newblock \emph{Preprint}, arXiv:2212.09058.

\bibitem[{Chen et~al.(2024{\natexlab{a}})Chen, Liang, Ma, Zheng, and Chen}]{chen2024eatselfsupervisedpretrainingefficient}
Wenxi Chen, Yuzhe Liang, Ziyang Ma, Zhisheng Zheng, and Xie Chen. 2024{\natexlab{a}}.
\newblock \href {https://arxiv.org/abs/2401.03497} {Eat: Self-supervised pre-training with efficient audio transformer}.
\newblock \emph{Preprint}, arXiv:2401.03497.

\bibitem[{Chen et~al.(2024{\natexlab{b}})Chen, Liang, Ma, Zheng, and Chen}]{ijcai2024p421}
Wenxi Chen, Yuzhe Liang, Ziyang Ma, Zhisheng Zheng, and Xie Chen. 2024{\natexlab{b}}.
\newblock \href {https://doi.org/10.24963/ijcai.2024/421} {Eat: Self-supervised pre-training with efficient audio transformer}.
\newblock In \emph{Proceedings of the Thirty-Third International Joint Conference on Artificial Intelligence, {IJCAI-24}}, pages 3807--3815. International Joint Conferences on Artificial Intelligence Organization.
\newblock Main Track.

\bibitem[{Cheng et~al.(2024)Cheng, Zhang, Zhang, Meng, Hong, Li, Wang, Wang, Yin, Zhao, and He}]{MAS-cheng2024exploringlargelanguagemodel}
Yuheng Cheng, Ceyao Zhang, Zhengwen Zhang, Xiangrui Meng, Sirui Hong, Wenhao Li, Zihao Wang, Zekai Wang, Feng Yin, Junhua Zhao, and Xiuqiang He. 2024.
\newblock \href {https://arxiv.org/abs/2401.03428} {Exploring large language model based intelligent agents: Definitions, methods, and prospects}.
\newblock \emph{Preprint}, arXiv:2401.03428.

\bibitem[{Chu et~al.(2024)Chu, Xu, Yang, Wei, Wei, Guo, Leng, Lv, He, Lin, Zhou, and Zhou}]{chu2024qwen2audiotechnicalreport}
Yunfei Chu, Jin Xu, Qian Yang, Haojie Wei, Xipin Wei, Zhifang Guo, Yichong Leng, Yuanjun Lv, Jinzheng He, Junyang Lin, Chang Zhou, and Jingren Zhou. 2024.
\newblock \href {https://arxiv.org/abs/2407.10759} {Qwen2-audio technical report}.
\newblock \emph{Preprint}, arXiv:2407.10759.

\bibitem[{Conia et~al.(2024)Conia, Lee, Li, Minhas, Potdar, and Li}]{conia-etal-2024-towards}
Simone Conia, Daniel Lee, Min Li, Umar~Farooq Minhas, Saloni Potdar, and Yunyao Li. 2024.
\newblock \href {https://doi.org/10.18653/v1/2024.emnlp-main.914} {Towards cross-cultural machine translation with retrieval-augmented generation from multilingual knowledge graphs}.
\newblock In \emph{Proceedings of the 2024 Conference on Empirical Methods in Natural Language Processing}, pages 16343--16360, Miami, Florida, USA. Association for Computational Linguistics.

\bibitem[{Ding et~al.(2023)Ding, Zhang, Amiri, Cao, Yang, Kaminski, Esselink, and Zhang}]{DS-Ding_2023}
Yan Ding, Xiaohan Zhang, Saeid Amiri, Nieqing Cao, Hao Yang, Andy Kaminski, Chad Esselink, and Shiqi Zhang. 2023.
\newblock \href {https://doi.org/10.1007/s10514-023-10133-5} {Integrating action knowledge and llms for task planning and situation handling in open worlds}.
\newblock \emph{Autonomous Robots}, 47(8):981–997.

\bibitem[{Ding et~al.(2024)Ding, Ren, Huang, Luo, and Han}]{ding2024pdfmvqadatasetmultimodalinformation}
Yihao Ding, Kaixuan Ren, Jiabin Huang, Siwen Luo, and Soyeon~Caren Han. 2024.
\newblock \href {https://arxiv.org/abs/2404.12720} {Pdf-mvqa: A dataset for multimodal information retrieval in pdf-based visual question answering}.
\newblock \emph{Preprint}, arXiv:2404.12720.

\bibitem[{et~al.(2024)}]{grattafiori2024llama3herdmodels}
Aaron~Grattafiori et~al. 2024.
\newblock \href {https://arxiv.org/abs/2407.21783} {The llama 3 herd of models}.
\newblock \emph{Preprint}, arXiv:2407.21783.

\bibitem[{et~al.(2021)}]{radford2021learningtransferablevisualmodels}
Alec~Radford et~al. 2021.
\newblock \href {https://arxiv.org/abs/2103.00020} {Learning transferable visual models from natural language supervision}.
\newblock \emph{Preprint}, arXiv:2103.00020.

\bibitem[{et~al.(2025)}]{villacueva2025cammtbenchmarkingculturallyaware}
Emilio Villa-Cueva et~al. 2025.
\newblock \href {https://arxiv.org/abs/2505.24456} {Cammt: Benchmarking culturally aware multimodal machine translation}.
\newblock \emph{Preprint}, arXiv:2505.24456.

\bibitem[{Freitag et~al.(2020)Freitag, Grangier, and Caswell}]{freitag-etal-2020-bleu}
Markus Freitag, David Grangier, and Isaac Caswell. 2020.
\newblock \href {https://doi.org/10.18653/v1/2020.emnlp-main.5} {{BLEU} might be guilty but references are not innocent}.
\newblock In \emph{Proceedings of the 2020 Conference on Empirical Methods in Natural Language Processing (EMNLP)}, pages 61--71, Online. Association for Computational Linguistics.

\bibitem[{Gaido et~al.(2024)Gaido, Papi, Negri, Cettolo, and Bentivogli}]{gaido-et-al-2024-sbaam}
Marco Gaido, Sara Papi, Matteo Negri, Mauro Cettolo, and Luisa Bentivogli. 2024.
\newblock {SBAAM! Eliminating Transcript Dependency in Automatic Subtitling}.
\newblock In \emph{Proceedings of the 62nd Annual Meeting of the Association for Computational Linguistics (Volume 1: Long Papers)}, Bangkok, Thailand.

\bibitem[{Gao et~al.(2023{\natexlab{a}})Gao, Wang, and Hou}]{gao2023designtranslationpromptschatgpt}
Yuan Gao, Ruili Wang, and Feng Hou. 2023{\natexlab{a}}.
\newblock \href {https://arxiv.org/abs/2304.02182} {How to design translation prompts for chatgpt: An empirical study}.
\newblock \emph{Preprint}, arXiv:2304.02182.

\bibitem[{Gao et~al.(2023{\natexlab{b}})Gao, Zhang, McLoughlin, and Yan}]{gao2023paraformerfastaccurateparallel}
Zhifu Gao, Shiliang Zhang, Ian McLoughlin, and Zhijie Yan. 2023{\natexlab{b}}.
\newblock \href {https://arxiv.org/abs/2206.08317} {Paraformer: Fast and accurate parallel transformer for non-autoregressive end-to-end speech recognition}.
\newblock \emph{Preprint}, arXiv:2206.08317.

\bibitem[{{Google}(2024)}]{GoogleTranslate}
{Google}. 2024.
\newblock Google translate.
\newblock \url{https://translate.google.com}.
\newblock Accessed: 2024-07-05.

\bibitem[{{Google DeepMind} and {Google Research}(2025)}]{gemini2.5flash2025}
{Google DeepMind} and {Google Research}. 2025.
\newblock \href {https://storage.googleapis.com/model-cards/documents/gemini-2.5-flash.pdf} {Gemini 2.5 flash}.
\newblock Model card, Google AI Studio / Vertex AI.
\newblock Enhanced multimodal large language model with extended capabilities in text, image, audio, and video.

\bibitem[{Guo et~al.(2024{\natexlab{a}})Guo, Zhang, Ma, Zhang, and Feng}]{guo2024agentsimtagentassistedsimultaneousmachine}
Shoutao Guo, Shaolei Zhang, Zhengrui Ma, Min Zhang, and Yang Feng. 2024{\natexlab{a}}.
\newblock \href {https://arxiv.org/abs/2406.06910} {Agent-simt: Agent-assisted simultaneous machine translation with large language models}.
\newblock \emph{Preprint}, arXiv:2406.06910.

\bibitem[{Guo et~al.(2024{\natexlab{b}})Guo, Chen, Wang, Chang, Pei, Chawla, Wiest, and Zhang}]{MAS-ijcai2024p890}
Taicheng Guo, Xiuying Chen, Yaqi Wang, Ruidi Chang, Shichao Pei, Nitesh~V. Chawla, Olaf Wiest, and Xiangliang Zhang. 2024{\natexlab{b}}.
\newblock \href {https://doi.org/10.24963/ijcai.2024/890} {Large language model based multi-agents: A survey of progress and challenges}.
\newblock In \emph{Proceedings of the Thirty-Third International Joint Conference on Artificial Intelligence, {IJCAI-24}}, pages 8048--8057. International Joint Conferences on Artificial Intelligence Organization.
\newblock Survey Track.

\bibitem[{He et~al.(2024)He, Liang, Jiao, Zhang, Yang, Wang, Tu, Shi, and Wang}]{he-etal-2024-exploring}
Zhiwei He, Tian Liang, Wenxiang Jiao, Zhuosheng Zhang, Yujiu Yang, Rui Wang, Zhaopeng Tu, Shuming Shi, and Xing Wang. 2024.
\newblock \href {https://doi.org/10.1162/tacl_a_00642} {Exploring human-like translation strategy with large language models}.
\newblock \emph{Transactions of the Association for Computational Linguistics}, 12:229--246.

\bibitem[{Hendy et~al.(2023)Hendy, Abdelrehim, Sharaf, Raunak, Gabr, Matsushita, Kim, Afify, and Awadalla}]{hendy2023goodgptmodelsmachine}
Amr Hendy, Mohamed Abdelrehim, Amr Sharaf, Vikas Raunak, Mohamed Gabr, Hitokazu Matsushita, Young~Jin Kim, Mohamed Afify, and Hany~Hassan Awadalla. 2023.
\newblock \href {https://arxiv.org/abs/2302.09210} {How good are gpt models at machine translation? a comprehensive evaluation}.
\newblock \emph{Preprint}, arXiv:2302.09210.

\bibitem[{Hu et~al.(2021)Hu, Bhowmick, Jang, Arvin, and Lanzon}]{MAS-9423979}
Junyan Hu, Parijat Bhowmick, Inmo Jang, Farshad Arvin, and Alexander Lanzon. 2021.
\newblock \href {https://doi.org/10.1109/TRO.2021.3071615} {A decentralized cluster formation containment framework for multirobot systems}.
\newblock \emph{IEEE Transactions on Robotics}, 37(6):1936--1955.

\bibitem[{Jiao et~al.(2023)Jiao, Wang, tse Huang, Wang, Shi, and Tu}]{jiao2023chatgptgoodtranslatoryes}
Wenxiang Jiao, Wenxuan Wang, Jen tse Huang, Xing Wang, Shuming Shi, and Zhaopeng Tu. 2023.
\newblock \href {https://arxiv.org/abs/2301.08745} {Is chatgpt a good translator? yes with gpt-4 as the engine}.
\newblock \emph{Preprint}, arXiv:2301.08745.

\bibitem[{Kang et~al.(2023)Kang, Huang, Peng, Zhu, Sun, Cheng, Wang, Huang, and Su}]{kang-etal-2023-bigvideo}
Liyan Kang, Luyang Huang, Ningxin Peng, Peihao Zhu, Zewei Sun, Shanbo Cheng, Mingxuan Wang, Degen Huang, and Jinsong Su. 2023.
\newblock \href {https://doi.org/10.18653/v1/2023.findings-acl.535} {{B}ig{V}ideo: A large-scale video subtitle translation dataset for multimodal machine translation}.
\newblock In \emph{Findings of the Association for Computational Linguistics: ACL 2023}, pages 8456--8473, Toronto, Canada. Association for Computational Linguistics.

\bibitem[{Kirillov et~al.(2023)Kirillov, Mintun, Ravi, Mao, Rolland, Gustafson, Xiao, Whitehead, Berg, Lo, Dollár, and Girshick}]{kirillov2023segment}
Alexander Kirillov, Eric Mintun, Nikhila Ravi, Hanzi Mao, Chloe Rolland, Laura Gustafson, Tete Xiao, Spencer Whitehead, Alexander~C. Berg, Wan-Yen Lo, Piotr Dollár, and Ross Girshick. 2023.
\newblock \href {https://arxiv.org/abs/2304.02643} {Segment anything}.
\newblock \emph{Preprint}, arXiv:2304.02643.

\bibitem[{Lan et~al.(2023)Lan, Yu, Li, Zhang, Luan, Wang, Huang, and Su}]{lan2023exploringbettertextimage}
Zhibin Lan, Jiawei Yu, Xiang Li, Wen Zhang, Jian Luan, Bin Wang, Degen Huang, and Jinsong Su. 2023.
\newblock \href {https://arxiv.org/abs/2305.17415} {Exploring better text image translation with multimodal codebook}.
\newblock \emph{Preprint}, arXiv:2305.17415.

\bibitem[{Li et~al.(2022)Li, Lv, Zhou, Zhou, Xiao, Ma, and Zhu}]{li-etal-2022-vision}
Bei Li, Chuanhao Lv, Zefan Zhou, Tao Zhou, Tong Xiao, Anxiang Ma, and JingBo Zhu. 2022.
\newblock \href {https://doi.org/10.18653/v1/2022.acl-long.438} {On vision features in multimodal machine translation}.
\newblock In \emph{Proceedings of the 60th Annual Meeting of the Association for Computational Linguistics (Volume 1: Long Papers)}, pages 6327--6337, Dublin, Ireland. Association for Computational Linguistics.

\bibitem[{Li et~al.(2023{\natexlab{a}})Li, Hammoud, Itani, Khizbullin, and Ghanem}]{li2023camel}
Guohao Li, Hasan Abed Al~Kader Hammoud, Hani Itani, Dmitrii Khizbullin, and Bernard Ghanem. 2023{\natexlab{a}}.
\newblock Camel: Communicative agents for "mind" exploration of large language model society.
\newblock In \emph{Thirty-seventh Conference on Neural Information Processing Systems}.

\bibitem[{Li et~al.(2023{\natexlab{b}})Li, Zhao, Yu, Song, Li, Yu, Li, Huang, and Li}]{tool-li2023apibankcomprehensivebenchmarktoolaugmented}
Minghao Li, Yingxiu Zhao, Bowen Yu, Feifan Song, Hangyu Li, Haiyang Yu, Zhoujun Li, Fei Huang, and Yongbin Li. 2023{\natexlab{b}}.
\newblock \href {https://arxiv.org/abs/2304.08244} {Api-bank: A comprehensive benchmark for tool-augmented llms}.
\newblock \emph{Preprint}, arXiv:2304.08244.

\bibitem[{Li et~al.(2023{\natexlab{c}})Li, Qiao, Wang, Wang, Jin, and Zha}]{decision-li2023semanticallyalignedtaskdecomposition}
Wenhao Li, Dan Qiao, Baoxiang Wang, Xiangfeng Wang, Bo~Jin, and Hongyuan Zha. 2023{\natexlab{c}}.
\newblock \href {https://arxiv.org/abs/2305.10865} {Semantically aligned task decomposition in multi-agent reinforcement learning}.
\newblock \emph{Preprint}, arXiv:2305.10865.

\bibitem[{Lin et~al.(2020)Lin, Meng, Su, Yin, Yang, Ge, Zhou, and Luo}]{Lin_2020}
Huan Lin, Fandong Meng, Jinsong Su, Yongjing Yin, Zhengyuan Yang, Yubin Ge, Jie Zhou, and Jiebo Luo. 2020.
\newblock \href {https://doi.org/10.1145/3394171.3413715} {Dynamic context-guided capsule network for multimodal machine translation}.
\newblock In \emph{Proceedings of the 28th ACM International Conference on Multimedia}, MM ’20. ACM.

\bibitem[{Liu(2022)}]{Liu_LlamaIndex_2022}
Jerry Liu. 2022.
\newblock \href {https://doi.org/10.5281/zenodo.1234} {{LlamaIndex}}.

\bibitem[{Long et~al.(2024)Long, Zeng, Meng, Ma, Zhang, Zhou, and Zhou}]{long2024generativemultimodalknowledgeretrieval}
Xinwei Long, Jiali Zeng, Fandong Meng, Zhiyuan Ma, Kaiyan Zhang, Bowen Zhou, and Jie Zhou. 2024.
\newblock \href {https://arxiv.org/abs/2401.08206} {Generative multi-modal knowledge retrieval with large language models}.
\newblock \emph{Preprint}, arXiv:2401.08206.

\bibitem[{Lu et~al.(2025)Lu, Sun, Zhao, Zhang, Song, and Yang}]{lu2025multimodalmachinetranslationvisual}
Chenyu Lu, Shiliang Sun, Jing Zhao, Nan Zhang, Tengfei Song, and Hao Yang. 2025.
\newblock \href {https://arxiv.org/abs/2505.19507} {Multimodal machine translation with visual scene graph pruning}.
\newblock \emph{Preprint}, arXiv:2505.19507.

\bibitem[{Lu et~al.(2024{\natexlab{a}})Lu, Song, Chang, Bian, Maiti, and Watanabe}]{lu2024syneslmunifiedapproachaudiovisual}
Yichen Lu, Jiaqi Song, Xuankai Chang, Hengwei Bian, Soumi Maiti, and Shinji Watanabe. 2024{\natexlab{a}}.
\newblock \href {https://arxiv.org/abs/2408.00624} {Syneslm: A unified approach for audio-visual speech recognition and translation via language model and synthetic data}.
\newblock \emph{Preprint}, arXiv:2408.00624.

\bibitem[{Lu et~al.(2024{\natexlab{b}})Lu, Song, Yang, and Watanabe}]{lu-etal-2024-fastadasp}
Yichen Lu, Jiaqi Song, Chao-Han~Huck Yang, and Shinji Watanabe. 2024{\natexlab{b}}.
\newblock \href {https://doi.org/10.18653/v1/2024.emnlp-industry.33} {{F}ast{A}da{SP}: Multitask-adapted efficient inference for large speech language model}.
\newblock In \emph{Proceedings of the 2024 Conference on Empirical Methods in Natural Language Processing: Industry Track}, pages 440--451, Miami, Florida, US. Association for Computational Linguistics.

\bibitem[{Lv et~al.(2025)Lv, Chen, Long, Fu, and Chen}]{lv2025topicvdtopicbaseddatasetvideoguided}
Jinze Lv, Jian Chen, Zi~Long, Xianghua Fu, and Yin Chen. 2025.
\newblock \href {https://arxiv.org/abs/2505.05714} {Topicvd: A topic-based dataset of video-guided multimodal machine translation for documentaries}.
\newblock \emph{Preprint}, arXiv:2505.05714.

\bibitem[{Ma et~al.(2024)Ma, Mi, Zeng, Yan, Wu, Lin, Zhang, and Wang}]{decision-NEURIPS2024_f0ebc318}
Weiyu Ma, Qirui Mi, Yongcheng Zeng, Xue Yan, Yuqiao Wu, Runji Lin, Haifeng Zhang, and Jun Wang. 2024.
\newblock \href {https://proceedings.neurips.cc/paper_files/paper/2024/file/f0ebc318e2df08360b2df559e81602e5-Paper-Conference.pdf} {Large language models play starcraft ii:benchmarks and a chain of summarization approach}.
\newblock In \emph{Advances in Neural Information Processing Systems}, volume~37, pages 133386--133442. Curran Associates, Inc.

\bibitem[{Ma et~al.(2023)Ma, Zheng, Ye, Li, Gao, Zhang, and Chen}]{ma2023emotion2vecselfsupervisedpretrainingspeech}
Ziyang Ma, Zhisheng Zheng, Jiaxin Ye, Jinchao Li, Zhifu Gao, Shiliang Zhang, and Xie Chen. 2023.
\newblock \href {https://arxiv.org/abs/2312.15185} {emotion2vec: Self-supervised pre-training for speech emotion representation}.
\newblock \emph{Preprint}, arXiv:2312.15185.

\bibitem[{Merx et~al.(2024)Merx, Mahmudi, Langford, de~Araujo, and Vylomova}]{merx2024lowresourcemachinetranslationretrievalaugmented}
Raphaël Merx, Aso Mahmudi, Katrina Langford, Leo~Alberto de~Araujo, and Ekaterina Vylomova. 2024.
\newblock \href {https://arxiv.org/abs/2404.04809} {Low-resource machine translation through retrieval-augmented llm prompting: A study on the mambai language}.
\newblock \emph{Preprint}, arXiv:2404.04809.

\bibitem[{OpenAI(2024)}]{openai2024gpt4ocard}
et~al. OpenAI. 2024.
\newblock \href {https://arxiv.org/abs/2410.21276} {Gpt-4o system card}.
\newblock \emph{Preprint}, arXiv:2410.21276.

\bibitem[{Peter et~al.(2024)Peter, Dang, Liu, Dominguez, and Lohia}]{peter2024multi}
Anishka Peter, Mai Dang, Michael Liu, Joaquin Dominguez, and Nibhrat Lohia. 2024.
\newblock Multi-agent translation team (matt): Enhancing low-resource language translation through multi-agent workflow.
\newblock \emph{SMU Data Science Review}, 8(3):3.

\bibitem[{Plaquet and Bredin(2023{\natexlab{a}})}]{Plaquet23}
Alexis Plaquet and Hervé Bredin. 2023{\natexlab{a}}.
\newblock {Powerset multi-class cross entropy loss for neural speaker diarization}.
\newblock In \emph{Proc. INTERSPEECH 2023}.

\bibitem[{Plaquet and Bredin(2023{\natexlab{b}})}]{Plaquet_2023}
Alexis Plaquet and Hervé Bredin. 2023{\natexlab{b}}.
\newblock \href {https://doi.org/10.21437/interspeech.2023-205} {Powerset multi-class cross entropy loss for neural speaker diarization}.
\newblock In \emph{INTERSPEECH 2023}, interspeech\_2023. ISCA.

\bibitem[{Qwen et~al.(2025)Qwen, :, and et~al.}]{qwen2025qwen25technicalreport}
Qwen, :, and An~Yang et~al. 2025.
\newblock \href {https://arxiv.org/abs/2412.15115} {Qwen2.5 technical report}.
\newblock \emph{Preprint}, arXiv:2412.15115.

\bibitem[{Radford et~al.(2022)Radford, Kim, Xu, Brockman, McLeavey, and Sutskever}]{radford2022robustspeechrecognitionlargescale}
Alec Radford, Jong~Wook Kim, Tao Xu, Greg Brockman, Christine McLeavey, and Ilya Sutskever. 2022.
\newblock \href {https://arxiv.org/abs/2212.04356} {Robust speech recognition via large-scale weak supervision}.
\newblock \emph{Preprint}, arXiv:2212.04356.

\bibitem[{Rei et~al.(2020)Rei, Stewart, Farinha, and Lavie}]{rei-etal-2020-comet}
Ricardo Rei, Craig Stewart, Ana~C Farinha, and Alon Lavie. 2020.
\newblock \href {https://doi.org/10.18653/v1/2020.emnlp-main.213} {{COMET}: A neural framework for {MT} evaluation}.
\newblock In \emph{Proceedings of the 2020 Conference on Empirical Methods in Natural Language Processing (EMNLP)}, pages 2685--2702, Online. Association for Computational Linguistics.

\bibitem[{Robinson et~al.(2023)Robinson, Ogayo, Mortensen, and Neubig}]{robinson-etal-2023-chatgpt}
Nathaniel Robinson, Perez Ogayo, David~R. Mortensen, and Graham Neubig. 2023.
\newblock \href {https://doi.org/10.18653/v1/2023.wmt-1.40} {{C}hat{GPT} {MT}: Competitive for high- (but not low-) resource languages}.
\newblock In \emph{Proceedings of the Eighth Conference on Machine Translation}, pages 392--418, Singapore. Association for Computational Linguistics.

\bibitem[{Ruan et~al.(2023)Ruan, Chen, Zhang, Xu, Bao, Du, Shi, Mao, Li, Zeng, and Zhao}]{tool-ruan2023tptulargelanguagemodelbased}
Jingqing Ruan, Yihong Chen, Bin Zhang, Zhiwei Xu, Tianpeng Bao, Guoqing Du, Shiwei Shi, Hangyu Mao, Ziyue Li, Xingyu Zeng, and Rui Zhao. 2023.
\newblock \href {https://arxiv.org/abs/2308.03427} {Tptu: Large language model-based ai agents for task planning and tool usage}.
\newblock \emph{Preprint}, arXiv:2308.03427.

\bibitem[{Sellam et~al.(2020)Sellam, Das, and Parikh}]{sellam2020bleurt}
Thibault Sellam, Dipanjan Das, and Ankur~P Parikh. 2020.
\newblock Bleurt: Learning robust metrics for text generation.
\newblock In \emph{Proceedings of ACL}.

\bibitem[{Shen et~al.(2024)Shen, Shao, Li, Lan, Liu, and Su}]{shen2024surveymultimodalmachinetranslation}
Huangjun Shen, Liangying Shao, Wenbo Li, Zhibin Lan, Zhanyu Liu, and Jinsong Su. 2024.
\newblock \href {https://arxiv.org/abs/2405.12669} {A survey on multi-modal machine translation: Tasks, methods and challenges}.
\newblock \emph{Preprint}, arXiv:2405.12669.

\bibitem[{Singhal et~al.(2023{\natexlab{a}})Singhal, Azizi, Tu, Mahdavi, Wei, Chung, Scales, Tanwani, Cole-Lewis, Pfohl, Payne, Seneviratne, Gamble, Kelly, Babiker, Sch{\"a}rli, Chowdhery, Mansfield, Demner-Fushman, Ag{\"u}era~y Arcas, Webster, Corrado, Matias, Chou, Gottweis, Tomasev, Liu, Rajkomar, Barral, Semturs, Karthikesalingam, and Natarajan}]{DS-naturecite-key}
Karan Singhal, Shekoofeh Azizi, Tao Tu, S.~Sara Mahdavi, Jason Wei, Hyung~Won Chung, Nathan Scales, Ajay Tanwani, Heather Cole-Lewis, Stephen Pfohl, Perry Payne, Martin Seneviratne, Paul Gamble, Chris Kelly, Abubakr Babiker, Nathanael Sch{\"a}rli, Aakanksha Chowdhery, Philip Mansfield, Dina Demner-Fushman, Blaise Ag{\"u}era~y Arcas, Dale Webster, Greg~S. Corrado, Yossi Matias, Katherine Chou, Juraj Gottweis, Nenad Tomasev, Yun Liu, Alvin Rajkomar, Joelle Barral, Christopher Semturs, Alan Karthikesalingam, and Vivek Natarajan. 2023{\natexlab{a}}.
\newblock \href {https://doi.org/10.1038/s41586-023-06291-2} {Large language models encode clinical knowledge}.
\newblock \emph{Nature}, 620(7972):172--180.

\bibitem[{Singhal et~al.(2023{\natexlab{b}})Singhal, Tu, Gottweis, Sayres, Wulczyn, Hou, Clark, Pfohl, Cole-Lewis, Neal, Schaekermann, Wang, Amin, Lachgar, Mansfield, Prakash, Green, Dominowska, y~Arcas, Tomasev, Liu, Wong, Semturs, Mahdavi, Barral, Webster, Corrado, Matias, Azizi, Karthikesalingam, and Natarajan}]{DS-singhal2023expertlevelmedicalquestionanswering}
Karan Singhal, Tao Tu, Juraj Gottweis, Rory Sayres, Ellery Wulczyn, Le~Hou, Kevin Clark, Stephen Pfohl, Heather Cole-Lewis, Darlene Neal, Mike Schaekermann, Amy Wang, Mohamed Amin, Sami Lachgar, Philip Mansfield, Sushant Prakash, Bradley Green, Ewa Dominowska, Blaise~Aguera y~Arcas, Nenad Tomasev, Yun Liu, Renee Wong, Christopher Semturs, S.~Sara Mahdavi, Joelle Barral, Dale Webster, Greg~S. Corrado, Yossi Matias, Shekoofeh Azizi, Alan Karthikesalingam, and Vivek Natarajan. 2023{\natexlab{b}}.
\newblock \href {https://arxiv.org/abs/2305.09617} {Towards expert-level medical question answering with large language models}.
\newblock \emph{Preprint}, arXiv:2305.09617.

\bibitem[{Sulubacak et~al.(2019)Sulubacak, Caglayan, Grönroos, Rouhe, Elliott, Specia, and Tiedemann}]{sulubacak2019multimodalmachinetranslationvisuals}
Umut Sulubacak, Ozan Caglayan, Stig-Arne Grönroos, Aku Rouhe, Desmond Elliott, Lucia Specia, and Jörg Tiedemann. 2019.
\newblock \href {https://arxiv.org/abs/1911.12798} {Multimodal machine translation through visuals and speech}.
\newblock \emph{Preprint}, arXiv:1911.12798.

\bibitem[{Tang et~al.(2024)Tang, Yu, Sun, Chen, Tan, Li, Lu, Ma, and Zhang}]{tang2024salmonngenerichearingabilities}
Changli Tang, Wenyi Yu, Guangzhi Sun, Xianzhao Chen, Tian Tan, Wei Li, Lu~Lu, Zejun Ma, and Chao Zhang. 2024.
\newblock \href {https://arxiv.org/abs/2310.13289} {Salmonn: Towards generic hearing abilities for large language models}.
\newblock \emph{Preprint}, arXiv:2310.13289.

\bibitem[{{Tavily AI}(2025)}]{tavily}
{Tavily AI}. 2025.
\newblock \href {https://www.tavily.com/} {Tavily}.
\newblock Accessed: 2025-07-05.

\bibitem[{Team and et~al.(2022)}]{nllbteam2022languageleftbehindscaling}
NLLB Team and et~al. 2022.
\newblock \href {https://arxiv.org/abs/2207.04672} {No language left behind: Scaling human-centered machine translation}.
\newblock \emph{Preprint}, arXiv:2207.04672.

\bibitem[{Tomar(2006)}]{tomar2006converting}
Suramya Tomar. 2006.
\newblock Converting video formats with ffmpeg.
\newblock \emph{Linux journal}, 2006(146):10.

\bibitem[{Wang et~al.(2024)Wang, Meng, Zhang, and Zhou}]{wang2024retrievalaugmentedmachinetranslationunstructured}
Jiaan Wang, Fandong Meng, Yingxue Zhang, and Jie Zhou. 2024.
\newblock \href {https://arxiv.org/abs/2412.04342} {Retrieval-augmented machine translation with unstructured knowledge}.
\newblock \emph{Preprint}, arXiv:2412.04342.

\bibitem[{Wang et~al.(2023)Wang, Liu, Zheng, Qi, Chen, Yang, Zhao, Wang, Song, and Huang}]{decision-wang2023avalonsgamethoughtsbattle}
Shenzhi Wang, Chang Liu, Zilong Zheng, Siyuan Qi, Shuo Chen, Qisen Yang, Andrew Zhao, Chaofei Wang, Shiji Song, and Gao Huang. 2023.
\newblock \href {https://arxiv.org/abs/2310.01320} {Avalon's game of thoughts: Battle against deception through recursive contemplation}.
\newblock \emph{Preprint}, arXiv:2310.01320.

\bibitem[{Wang et~al.(2025{\natexlab{a}})Wang, Zeng, Liu, Wong, Meng, Zhou, and Zhang}]{wang2025deltaonlinedocumentleveltranslation}
Yutong Wang, Jiali Zeng, Xuebo Liu, Derek~F. Wong, Fandong Meng, Jie Zhou, and Min Zhang. 2025{\natexlab{a}}.
\newblock \href {https://arxiv.org/abs/2410.08143} {Delta: An online document-level translation agent based on multi-level memory}.
\newblock \emph{Preprint}, arXiv:2410.08143.

\bibitem[{Wang et~al.(2025{\natexlab{b}})Wang, Zeng, Liu, Wong, Meng, Zhou, and Zhang}]{MAS-mt-wang2025deltaonlinedocumentleveltranslation}
Yutong Wang, Jiali Zeng, Xuebo Liu, Derek~F. Wong, Fandong Meng, Jie Zhou, and Min Zhang. 2025{\natexlab{b}}.
\newblock \href {https://arxiv.org/abs/2410.08143} {Delta: An online document-level translation agent based on multi-level memory}.
\newblock \emph{Preprint}, arXiv:2410.08143.

\bibitem[{Weifeng2333(2024)}]{videocaptioner2024}
Weifeng2333. 2024.
\newblock Videocaptioner: An open-source cascaded system for video subtitling.
\newblock \url{https://github.com/WEIFENG2333/VideoCaptioner}.
\newblock Accessed: 2025-07-04.

\bibitem[{Wilken et~al.(2022)Wilken, Georgakopoulou, and Matusov}]{wilken-etal-2022-suber}
Patrick Wilken, Panayota Georgakopoulou, and Evgeny Matusov. 2022.
\newblock \href {https://doi.org/10.18653/v1/2022.iwslt-1.1} {{S}ub{ER} - a metric for automatic evaluation of subtitle quality}.
\newblock In \emph{Proceedings of the 19th International Conference on Spoken Language Translation (IWSLT 2022)}, pages 1--10, Dublin, Ireland (in-person and online). Association for Computational Linguistics.

\bibitem[{Wu et~al.(2024{\natexlab{a}})Wu, Xu, and Wang}]{wu-etal-2024-transagents}
Minghao Wu, Jiahao Xu, and Longyue Wang. 2024{\natexlab{a}}.
\newblock \href {https://doi.org/10.18653/v1/2024.emnlp-demo.14} {{T}rans{A}gents: Build your translation company with language agents}.
\newblock In \emph{Proceedings of the 2024 Conference on Empirical Methods in Natural Language Processing: System Demonstrations}, pages 131--141, Miami, Florida, USA. Association for Computational Linguistics.

\bibitem[{Wu et~al.(2023)Wu, Bansal, Zhang, Wu, Li, Zhu, Jiang, Zhang, Zhang, Liu, Awadallah, White, Burger, and Wang}]{tool-wu2023autogenenablingnextgenllm}
Qingyun Wu, Gagan Bansal, Jieyu Zhang, Yiran Wu, Beibin Li, Erkang Zhu, Li~Jiang, Xiaoyun Zhang, Shaokun Zhang, Jiale Liu, Ahmed~Hassan Awadallah, Ryen~W White, Doug Burger, and Chi Wang. 2023.
\newblock \href {https://arxiv.org/abs/2308.08155} {Autogen: Enabling next-gen llm applications via multi-agent conversation}.
\newblock \emph{Preprint}, arXiv:2308.08155.

\bibitem[{Wu et~al.(2024{\natexlab{b}})Wu, Lu, Peng, Wang, Song, and Watanabe}]{wu2024enhancingaudiovisualspeechrecognition}
Yihan Wu, Yichen Lu, Yifan Peng, Xihua Wang, Ruihua Song, and Shinji Watanabe. 2024{\natexlab{b}}.
\newblock \href {https://arxiv.org/abs/2412.19005} {Enhancing audiovisual speech recognition through bifocal preference optimization}.
\newblock \emph{Preprint}, arXiv:2412.19005.

\bibitem[{Wu et~al.(2016)Wu, Schuster, Chen, Le, Norouzi, Macherey, Krikun, Cao, Gao, Macherey, Klingner, Shah, Johnson, Liu, Łukasz Kaiser, Gouws, Kato, Kudo, Kazawa, Stevens, Kurian, Patil, Wang, Young, Smith, Riesa, Rudnick, Vinyals, Corrado, Hughes, and Dean}]{wu2016googlesneuralmachinetranslation}
Yonghui Wu, Mike Schuster, Zhifeng Chen, Quoc~V. Le, Mohammad Norouzi, Wolfgang Macherey, Maxim Krikun, Yuan Cao, Qin Gao, Klaus Macherey, Jeff Klingner, Apurva Shah, Melvin Johnson, Xiaobing Liu, Łukasz Kaiser, Stephan Gouws, Yoshikiyo Kato, Taku Kudo, Hideto Kazawa, Keith Stevens, George Kurian, Nishant Patil, Wei Wang, Cliff Young, Jason Smith, Jason Riesa, Alex Rudnick, Oriol Vinyals, Greg Corrado, Macduff Hughes, and Jeffrey Dean. 2016.
\newblock \href {https://arxiv.org/abs/1609.08144} {Google's neural machine translation system: Bridging the gap between human and machine translation}.
\newblock \emph{Preprint}, arXiv:1609.08144.

\bibitem[{Wu et~al.(2024{\natexlab{c}})Wu, Chen, Zhang, Hui, Nezhurina, Berg-Kirkpatrick, and Dubnov}]{wu2024largescalecontrastivelanguageaudiopretraining}
Yusong Wu, Ke~Chen, Tianyu Zhang, Yuchen Hui, Marianna Nezhurina, Taylor Berg-Kirkpatrick, and Shlomo Dubnov. 2024{\natexlab{c}}.
\newblock \href {https://arxiv.org/abs/2211.06687} {Large-scale contrastive language-audio pretraining with feature fusion and keyword-to-caption augmentation}.
\newblock \emph{Preprint}, arXiv:2211.06687.

\bibitem[{Xu et~al.(2024{\natexlab{a}})Xu, Sharaf, Chen, Tan, Shen, Durme, Murray, and Kim}]{xu2024contrastivepreferenceoptimizationpushing}
Haoran Xu, Amr Sharaf, Yunmo Chen, Weiting Tan, Lingfeng Shen, Benjamin~Van Durme, Kenton Murray, and Young~Jin Kim. 2024{\natexlab{a}}.
\newblock \href {https://arxiv.org/abs/2401.08417} {Contrastive preference optimization: Pushing the boundaries of llm performance in machine translation}.
\newblock \emph{Preprint}, arXiv:2401.08417.

\bibitem[{Xu et~al.(2025)Xu, Guo, He, Hu, He, Bai, Chen, Wang, Fan, Dang, Zhang, Wang, Chu, and Lin}]{xu2025qwen25omnitechnicalreport}
Jin Xu, Zhifang Guo, Jinzheng He, Hangrui Hu, Ting He, Shuai Bai, Keqin Chen, Jialin Wang, Yang Fan, Kai Dang, Bin Zhang, Xiong Wang, Yunfei Chu, and Junyang Lin. 2025.
\newblock \href {https://arxiv.org/abs/2503.20215} {Qwen2.5-omni technical report}.
\newblock \emph{Preprint}, arXiv:2503.20215.

\bibitem[{Xu et~al.(2018)Xu, Madotto, Wu, Park, and Fung}]{xu2018emo2veclearninggeneralizedemotion}
Peng Xu, Andrea Madotto, Chien-Sheng Wu, Ji~Ho Park, and Pascale Fung. 2018.
\newblock \href {https://arxiv.org/abs/1809.04505} {Emo2vec: Learning generalized emotion representation by multi-task training}.
\newblock \emph{Preprint}, arXiv:1809.04505.

\bibitem[{Xu et~al.(2024{\natexlab{b}})Xu, Yu, Fang, Wang, and Wu}]{decision-xu2024languageagentsreinforcementlearning}
Zelai Xu, Chao Yu, Fei Fang, Yu~Wang, and Yi~Wu. 2024{\natexlab{b}}.
\newblock \href {https://arxiv.org/abs/2310.18940} {Language agents with reinforcement learning for strategic play in the werewolf game}.
\newblock \emph{Preprint}, arXiv:2310.18940.

\bibitem[{Zhai(2024)}]{zhai2024selfadaptivemultimodalretrievalaugmentedgeneration}
Wenjia Zhai. 2024.
\newblock \href {https://arxiv.org/abs/2410.11321} {Self-adaptive multimodal retrieval-augmented generation}.
\newblock \emph{Preprint}, arXiv:2410.11321.

\bibitem[{Zhai et~al.(2023)Zhai, Mustafa, Kolesnikov, and Beyer}]{zhai2023sigmoidlosslanguageimage}
Xiaohua Zhai, Basil Mustafa, Alexander Kolesnikov, and Lucas Beyer. 2023.
\newblock \href {https://arxiv.org/abs/2303.15343} {Sigmoid loss for language image pre-training}.
\newblock \emph{Preprint}, arXiv:2303.15343.

\bibitem[{Zhu et~al.(2024)Zhu, Liu, Dong, Xu, Huang, Kong, Chen, and Li}]{zhu2024multilingualmachinetranslationlarge}
Wenhao Zhu, Hongyi Liu, Qingxiu Dong, Jingjing Xu, Shujian Huang, Lingpeng Kong, Jiajun Chen, and Lei Li. 2024.
\newblock \href {https://arxiv.org/abs/2304.04675} {Multilingual machine translation with large language models: Empirical results and analysis}.
\newblock \emph{Preprint}, arXiv:2304.04675.

\bibitem[{Zuo et~al.(2023)Zuo, Li, Lv, Zheng, Xiao, and Zhu}]{zuo-etal-2023-incorporating}
Yuxin Zuo, Bei Li, Chuanhao Lv, Tong Zheng, Tong Xiao, and JingBo Zhu. 2023.
\newblock \href {https://doi.org/10.18653/v1/2023.findings-emnlp.978} {Incorporating probing signals into multimodal machine translation via visual question-answering pairs}.
\newblock In \emph{Findings of the Association for Computational Linguistics: EMNLP 2023}, pages 14689--14701, Singapore. Association for Computational Linguistics.

\end{thebibliography}

\clearpage

\appendix
\label{sec:appendix}
\onecolumn
\section{ViDove Details}
\subsection{Multi-agent Translation System}
\vspace{2em}
\begin{table*}[h]
\centering
\resizebox{0.99\textwidth}{4cm}{\begin{tabular}{@{}c|c|c@{}}
\toprule
\multirow{3}{*}{Auditory Agent}                         & SpeechLM & \begin{tabular}[c]{@{}c@{}}SALMONN~\cite{tang2024salmonngenerichearingabilities}, \\ Gemini-2.5-flash~\cite{gemini2.5flash2025}\\ Qwen2-Audio~\cite{chu2024qwen2audiotechnicalreport}, Qwen-2.5-Omni~\cite{xu2025qwen25omnitechnicalreport},\end{tabular} \\ \cmidrule(l){2-3} 
                                                        & ASR      & \begin{tabular}[c]{@{}c@{}}Whisper-series~\cite{radford2022robustspeechrecognitionlargescale} \\ Paraformer\cite{gao2023paraformerfastaccurateparallel} \end{tabular}              \\ \cmidrule(l){2-3} 
                                                        & Others   & \begin{tabular}[c]{@{}c@{}}Emo2Vec\cite{xu2018emo2veclearninggeneralizedemotion}, \\ BEATs\cite{chen2022beatsaudiopretrainingacoustic}, EATs\cite{ijcai2024p421}\, CLAP\cite{wu2024largescalecontrastivelanguageaudiopretraining}\\ Pyannote\cite{Plaquet_2023}\end{tabular}              \\ \midrule
\multirow{2}{*}{Visual Agent}                           & VLMs    & \begin{tabular}[c]{@{}c@{}}GPT4-o\cite{openai2024gpt4ocard}, \\ Qwen2.5-VL\cite{Qwen2.5-VL}\end{tabular}                        \\ \cmidrule(l){2-3} 
                                                        & Others   & \begin{tabular}[c]{@{}c@{}}CLIP\cite{radford2021learningtransferablevisualmodels}, SigCLIP\cite{zhai2023sigmoidlosslanguageimage}, \\ SegAnything\cite{kirillov2023segment}\end{tabular}                         \\ \midrule
\multirow{2}{*}{Translation Agent \& Post-editing Team} & LLMs     & GPT-series\cite{openai2024gpt4ocard}, LLaMA-series\cite{grattafiori2024llama3herdmodels}, Qwen-series\cite{qwen2025qwen25technicalreport}                                                         \\ \cmidrule(l){2-3} 
                                                        & Others   & Google-translate\cite{GoogleTranslate}, NLLB\cite{nllbteam2022languageleftbehindscaling}                                                                        \\ \midrule
\multirow{2}{*}{Memory System}                          & Web      & Tavily\cite{tavily}                                                                                   \\ \cmidrule(l){2-3} 
                                                        & Local    & Llama-index\cite{Liu_LlamaIndex_2022}                                                                                   \\ \midrule
Other Tools                                             & -        & FFmpeg\cite{tomar2006converting}                                                                                        \\ \midrule
Evaluation Metrics & - & \makecell[c]{
BLEU~\cite{freitag-etal-2020-bleu}, COMET~\cite{rei-etal-2020-comet},
BLEURT~\cite{sellam2020bleurt},\\
SubER~\cite{wilken-etal-2022-suber}, SubSONAR~\cite{sulubacak2019multimodalmachinetranslationvisuals}
} \\
\bottomrule
\end{tabular}}
\caption{Base models and tools support by ViDove}
\label{tab:tools}
\end{table*}
\vspace{8em}
\begin{algorithm}[H]
\caption{Multi-agent Translation System}
\begin{algorithmic}
\Require Transcript chunk $T_i$, Memory $\mathcal{M}=\{\mathcal{M}^s,\mathcal{M}^l\}$, LLM Translator $\mathcal{L}_t$, Proofreader $\mathcal{L}_{pr}$, Editor $\mathcal{L}_{ed}$
\Ensure Translated transcript chunk $T^*_i$, updated short-term memory $\mathcal{M}^s_{history}$, and translation prompt $p_{translation}$
\State $translation\_history_{i, i-5} \leftarrow
\textbf{retrieve}
(\mathcal{M}^s_{history}, T_i)
$
\State $context_i \leftarrow \textbf{retrieve}(\mathcal{M}^s_{context})$
\State $domain\_guide \leftarrow \textbf{query}(\mathcal{M}^l_{domain}, T_i)$
\State $p_{translation} \leftarrow (p_{translation},context_i,domain\_guide)$
\State $T^*_i \leftarrow \mathcal{L}_t(T_i,p_{translation})$
\State $T^*_{i,pr} \leftarrow \mathcal{L}_{pr}(T^*_i, \mathcal{M}^s, \mathcal{M}^l)$ \Comment{Proofreader checks grammar, style, terminology}
\State $T^*_{i,ed} \leftarrow \mathcal{L}_{ed}(T^*_{i,pr}, \mathcal{M}^s, \mathcal{M}^l)$ \Comment{Editor ensures logical and contextual consistency}
\State $\mathcal{M}^s_{history} \leftarrow (\mathcal{M}^s, T_i, T^*_{i,ed})$
\State \Return $(T^*_{i,ed}, \mathcal{M}^s_{history})$
\end{algorithmic}
\label{algo-mt}
\caption{Multi-agent Translation Pipeline}
\end{algorithm}

\newpage

\subsubsection{Proofreader Agent}
\label{appendix：proofreader}

To further demonstrate the role of the proofreader agent in our pipeline, we present a series of log messages captured during a real session. The proofreader monitors intermediate translations and provides suggestions to correct terminology, sentence structure, and domain-specific references. We have provided examples(\ref{tab:proofreader_log_blocks}) illustrating its intervention.

\begin{CJK}{UTF8}{gbsn}
\begin{table}[H]
\renewcommand{\arraystretch}{1.35}
\centering
\small
\begin{tabular}{|p{14cm}|}
\hline
\texttt{[Segment 130] PASS} \\
\hline
\texttt{[Segment 131] The term "pilum" in the source text seems to be a mistake or unclear.} \\
\texttt{\hspace{2em}It might be intended to refer to "pylon" based on the term context provided.} \\
\texttt{\hspace{2em}Consider verifying this with the editor.} \\
\hline
\texttt{[Segment 132] The translation of "sporter" as "孢子" is incorrect.} \\
\texttt{\hspace{2em}Based on the term context, "sporter" might be a misinterpretation of "spore crawler,"} \\
\texttt{\hspace{2em}which should be translated as "孢子爬虫."} \\
\texttt{\hspace{2em}Verify with the editor if "sporter" is indeed meant to be "spore crawler."} \\
\hline
\texttt{[Segment 133] The translation is missing a verb or context to make it a complete sentence.} \\
\texttt{\hspace{2em}Consider adding context or a verb to improve fluency, such as} \\
\texttt{\hspace{2em}"显然现在是Nagra的时刻，伙计。"} \\
\hline
\texttt{[Segment 134] The translation of "Spire" as "空军基地" is incorrect.} \\
\texttt{\hspace{2em}According to the term context, "Spire" should be translated as "飞龙塔."} \\
\texttt{\hspace{2em}Adjust the translation to reflect this terminology.} \\
\hline
\end{tabular}
\caption{proofreader log}
\label{tab:proofreader_log_blocks}
\end{table}
\end{CJK}

The proofreader agent checks both the source and the translation, considering context information, which allows it to identify potential misinterpretations and suggest targeted corrections. In the above use case, it successfully detects anomalies in Segments 131 and 132. Segment 134 also showcases its ability to correct common terminology errors based on domain knowledge.

\subsection{ViDove Demo Page}
\FloatBarrier

\begin{figure*}[htbp]
\centering
\includegraphics[width=0.99\textwidth]{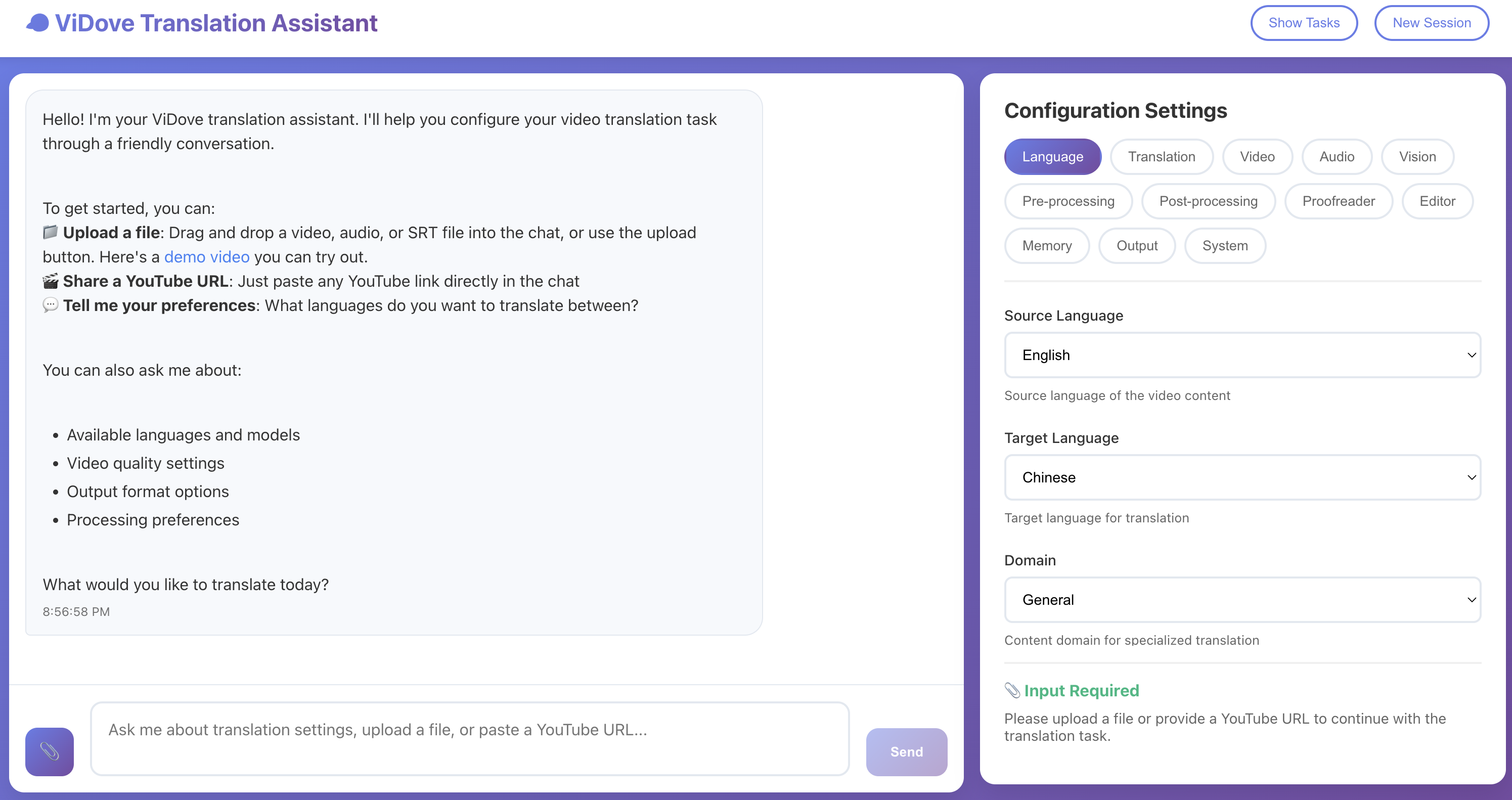} \\
\caption{\textbf{User interface of ViDove} }
\label{fig:demo}
\end{figure*}

\subsection{Prompt for ViDove Agents}

\vspace{2em}

\label{appendix:vidovePrompt}
\begin{tcolorbox}[breakable, fontupper=\small, title={ViDove Translator Agent}, after skip=\baselineskip]
\say{
You are a professional translator. your job is to translate texts in domain of \{domain\} from \{source language\} to \{target language\}
    
you will be provided with a segment in source language parsed by line, where your translation text should keep the original meaning and the number of lines.

Keep every \textbackslash n in the translated text in the corresponding place, and make sure to keep the same number of lines in the translated text.

You must break the translated sentence into multiple lines accordingly if original text breaks a complete sentence into different lines.

You should only output the translated text line by line without any other notation.

You current task is to translate the script in the domain of \{domain\} from \{source language\} to \{target language\}

Here are some supporting information including previous translation history, context documenting, supporting documents from internet and video clips description that might help you translate the text.
Please refer to them if necessary.
Previous translation history: \textbackslash n

\{Translation Histories\}

if you detect any word is in the following context, please use it as a reference for current translation

\{Context documents retrieved from knowledge base\}

Here are some supporting documents that might help you translate the text, refer to them if necessary.:

\{ supporting documents from web search \}

Here are some descriptions of video clips that might help you translate the text, refer to them if necessary. :

\{video clips descriptions\}

Now please translate the following text from \{source language\} to \{target language\}

\{{text to be translated}\}

Your translation:
}
\end{tcolorbox}
\vspace{2em}
\begin{tcolorbox}[breakable, fontupper=\small, title={ViDove Editor Agent}, after skip=\baselineskip]
\say{
You are an Editor ensuring overall translation quality and coherence,
aligning the translation with the original video content in domain \{domain\}, you must ensure the term and style are aligned with the domain's language.
        
Segment index: \{idx\}
Source text:
\{source\}

Translated text:
\{translation\}

Here is a provided suggestion for each segment, which may or may not useful for your revision, you may use the suggestion only if necessary (for example, term correctness).
Note that the suggestion may not be accurate, the proofreader has less information comparing to you, so you need to double check before making revision.
The proofreader may return "UNCLEAR" if they are not sure about the translation, they will specify the location and you need to check with other information provided to you to solve for unclear.
If there is no suggestions, you may ignore this part, but still check with other modality context and long-term memory for correctness and coherence.
Suggestion:

\{suggestion if suggestion else "No suggestion provided."\}

Your edit will also follow the following instruction if provided:
User instruction:
\{user instruction if user introduction else "No user instruction provided."\}
                
--- Multimodal Context (Short-Term Memory) ---
Visual cues:
You may use visual cues from the video to improve translation or make corrections, the source text might not be accurate, you need to check with the video context if provided:
\{visual context\}

Audio cues:
\{audio context\}

Translation context:
You will be provided with the previous and next 5 segments' translations, which may help you understand the context and make corrections:
Previous translation history (up to 5 segments):
\{Previous translation history\}
Past translation history (up to 5 segments):
\{Past translation history\}

--- Long-Term Memory ---
Long-term memory provides broader context and domain-specific knowledge, you may use it to improve translation or make corrections:
\{long term memory\}

Notice:
1. Corrections or adjustments to better align text with the video context.
2. Suggestions for improving coherence across segments.
3. Logical consistency and any broader context adjustments.
4. Ensure the translation is accurate and aligned with the domain \{domain\}.
5. Ensure translation is smooth and fluent across segments.
6. To ensure the fluency in \{target language\}, you do not have to ensure translation be word by word accurate, but be sure to convey the same information.

--- Important ---
Directly return the revised content only.
}
\end{tcolorbox}
\newpage
\begin{tcolorbox}[breakable, fontupper=\small, title={ViDove Proofreader Agent}, after skip=\baselineskip]
\say{
You are a translation proofreader. Below are \{number of segments\} subtitle segments.
Some are full sentences, some are fragments. Give **specific advice** for each one, 
but do not treat each segment separately you need information across segment.

Return suggestions in this format:
Segment 0: [your comment here]
Segment 1: [your comment here]
...

DO NOT return JSON. DO NOT rewrite the translation. Just return suggestion texts.

---
\{segments\}

**Short-term memory:**
\{short term memory\}

**Term context:**
\{local context\}

**Web memory context:**
\{web search context\}

Focus on:
1. Translation accuracy while sticking to domain \{domain\} (missing or incorrect meanings)
2. Fluency (grammar, spelling, repetition. Only if it affects understanding) and ensure the translation is smooth and fluent across segments.
3. Terminology (Use term context to edit idioms, ensure every sentence is translated into domain-specific language)
4. If you have no suggestions, return "PASS" for that segment.
5. Source text isn't 100\% accurate. If you have doubt about the source text, return "UNCLEAR" and specify the location, editor will check the issue.
6. Only make suggestions if you believe revision is necessary.
}
\end{tcolorbox}

\subsection{Translation Sample}
\FloatBarrier
\begin{figure*}[htbp]
\centering
\includegraphics[width=0.8\textwidth]{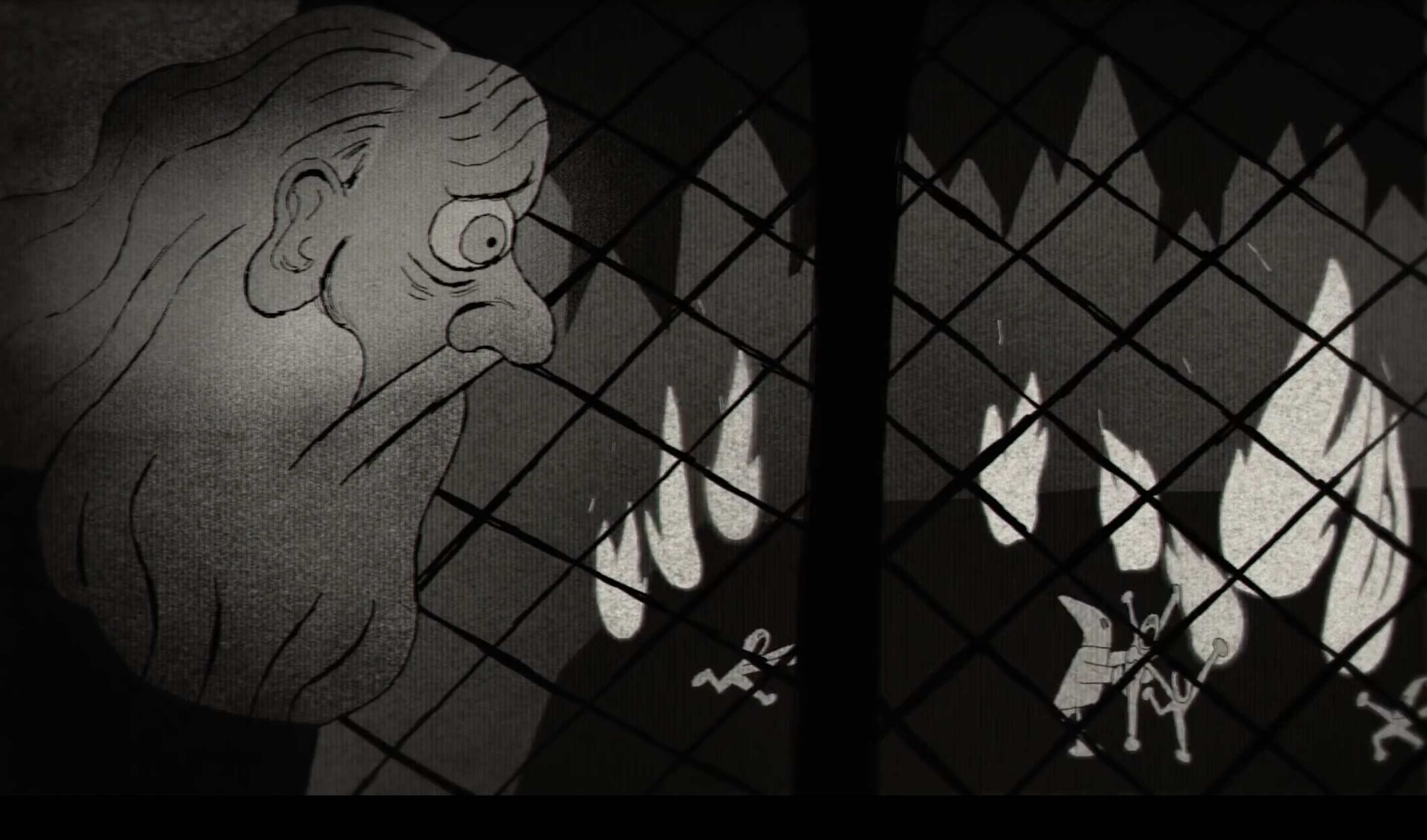} \\
\caption{\textbf{Vidove Video Cues Summarization for Translation Sample:} The image portrays a surreal and dramatic illustration featuring a large, exaggerated face of a bearded man on the left, expressing concern or thoughtfulness. On the right, a dark, patterned background showcases large white flames, with small skeletal figures interacting with them in the foreground. The scene is rendered in a monochromatic color scheme.}
\label{fig:demo}
\end{figure*}

\FloatBarrier
\begin{CJK}{UTF8}{gbsn}
\begin{table}[htbp]
\renewcommand{\arraystretch}{1.2}
\centering
\begin{tabular}{p{3cm} p{4cm}}
\hline
\textsc{Original Text} & What happens when the devil walks among us? \\
\hline
\textsc{Ground Truth}  & 若魔鬼化身凡人混迹其中，世界会变成什么样? \\
\hline
\textsc{Vidove} & 当魔鬼行走在人间时\textcolor{blue}{会发生什么？} \\
\hline
\textsc{VideoCaptioner} & 当魔鬼\textcolor{red}{\textsc{在我们中间行走会发生什么?}} \\
\hline
\end{tabular}
\caption{Case study for translation quality. \textcolor{blue}{Blue} highlights translation deviations to \textsc{Vidove}, and \textcolor{red}{red} highlights deviations to \textsc{VideoCaptioner}.}
\label{tab:dovebench}
\end{table}
\end{CJK}

\section{DoveBench}
\label{appendix:dovebench}





\subsection{DoveBench Stats}
DoveBench is a benchmark dataset designed to evaluate video translation and subtitling models. It contains a total of 50 videos, amounting to 17.23 hours of content and featuring 16,968 subtitle entries. The dataset's total text includes 189,157 words.

The dataset is composed of two distinct categories sourced from fan sub groups:
\begin{itemize}
    \item \textbf{CS}: This category includes 23 Counter-Strike related videos from the "fazeclan galaxy archive" fan sub group. The videos in this section have an average duration of approximately 13 minutes (777.7 seconds).
    \item \textbf{SC2}: This category consists of 27 videos in the StarCraft 2 domain from the "StarPigeon Fan sub group". These videos are generally longer, with an average duration of over 27 minutes (1635.3 seconds).
\end{itemize}

A key feature of DoveBench is the inclusion of detailed ground truth, which is essential for evaluation. We provide human-annotated ground-truth subtitles for all videos. The dataset's comprehensive statistics on character counts, word counts, duration, and subtitle distribution make it a valuable resource for assessing the performance of video translation systems.

\begin{CJK}{UTF8}{gbsn}
\begin{table}[htbp]
\renewcommand{\arraystretch}{1.2}
\centering
\resizebox{\textwidth}{!}{
\begin{tabular}{p{4.5cm} | c | c || p{5.5cm} | c}
\multicolumn{3}{c||}{\textbf{Category Statistics}} & \multicolumn{2}{c}{\textbf{Overall Dataset Statistics}} \\
\hline
\textbf{Statistic} & \textbf{CS} & \textbf{SC2} & \textbf{Statistic} & \textbf{Overall} \\
\hline
 &  &  & \textsc{Number of Videos} & 50 \\
\textsc{Number of Videos} & 23 & 27 & \textsc{Total Duration} & 17.23 hours \\
 &  &  & \textsc{Average Duration} & 20.68 minutes \\
\textsc{Total Duration} & 4.97 h (298.1 min) & 12.27 h (735.9 min)  & \textsc{Total Subtitle Lines} & 16,968 \\
 &  &  & \textsc{Avg. Subtitle Lines per Video} & 346.3 \\
textsc{Average Duration} & 12.96 min (777.7 s) & 27.26 min (1635.3 s)  & \textsc{Total Words} & 189,157 \\
 & &  & \textsc{Avg. Words per Video} & 3,860 \\
\hline
\end{tabular}
}
\caption{Detailed statistics of the DoveBench dataset}
\label{tab:dovebench_stats_detailed}
\end{table}
\end{CJK}

\section{Baseline Systems Details}
\subsection{Gemini Prompt for DoveBench}
\label{appendix:gemini}

\begin{tcolorbox}[breakable, fontupper=\small, title={Gemini Prompt (English Version)}, after skip=\baselineskip]
\say{
You are a professional transcription and translation assistant.

Please transcribe this audio/video file and translate it into Simplified Chinese. Carefully follow the instructions below:

1. Each segment should:
- Contain a natural sentence or phrase in Simplified Chinese, not too long.
- Have a valid start and end time in the format `h:mm:ss,ms` (e.g., "0:00:01,229").
- Ensure the start time is less than the end time, and that each segment’s start time equals the previous segment’s end time (no overlap or gap).
- If uncertain, round timestamps to the nearest 10 milliseconds.

2. Translation Guidelines:
- First, accurately understand the original audio content.
- Translate into natural, fluent Simplified Chinese.
- Retain the meaning and tone of the original speech.
- Keep proper nouns and technical terms accurate.
- Maintain sentence boundaries suitable for subtitle readability.

3. Notes:
- Proper nouns and technical terms — remain accurate.
- Sentence boundaries — avoid breaking at unnatural pauses.
- Chinese grammar and natural fluency.

Please provide the transcription and translation in the specified structured format.

The output language must be Chinese.
}
\end{tcolorbox}

\subsection{Qwen-2.5-Omni}
\label{appendix:qwen}

\subsubsection{Prompts for DoveBench and BigVideo}

\begin{CJK}{UTF8}{gbsn}
\begin{tcolorbox}[breakable, fontupper=\small, title={Qwen 2.5 Omni Prompt (Chinese Version)}, after skip=\baselineskip]
\say{
翻译提供的视频中的说话内容到中文。只需要输出翻译内容原文，不要输出任何解释。}
\end{tcolorbox}
\end{CJK}

\subsubsection{Prompt Design and Video Processing Strategy}

This section outlines the strategies employed for prompt design and video processing to optimize Qwen 2.5 Omni's performance on the DoveBench and BigVideo datasets.

\textbf{Prompt Language: }Given the superior performance of Chinese prompts in enhancing Qwen 2.5 Omni's instruction-following capabilities during preliminary evaluations, all experiments on both the DoveBench and BigVideo datasets exclusively utilized Chinese prompts. 

\textbf{Prompt Complexity: }Specifically, this approach was adopted because initial trials with prompts designed similarly to those employed for Gemini demonstrated that Qwen exhibited a deficiency in instruction following when presented with complex instructions. This observation led to the selection of the aforementioned  prompts, as they consistently yielded superior results.

\textbf{Video Processing Strategy: }Furthermore, due to Qwen 2.5 Omni's constrained contextual understanding when processing video data, most videos within the datasets—even those only one to two minutes in duration—could not be processed directly. To address this limitation, videos were manually segmented into approximately ten-second clips. Each segment was then individually fed into the model, and the processed outputs for each segment were subsequently concatenated to form a complete SRT file.

\end{document}